\documentclass[lettersize,journal]{IEEEtran}
\usepackage{amsmath,amsfonts}
\usepackage{algorithm}
\usepackage{algpseudocode}
\usepackage{array}
\usepackage{textcomp}
\usepackage{stfloats}
\usepackage{url}
\usepackage{verbatim}
\usepackage{graphicx}
\usepackage{cite}
\usepackage{threeparttable}
\usepackage{makecell}
\usepackage{multirow}
\usepackage{footnote}
\usepackage{tablefootnote}
\usepackage{float}
\usepackage[numbers]{natbib}
\usepackage{colortbl}
\usepackage{array}
\usepackage{caption}
\usepackage{booktabs}
\usepackage{ragged2e}
\usepackage{setspace}
\usepackage{enumitem}
\usepackage{subcaption}
\usepackage{float}
\usepackage{bbm}
\usepackage[table]{xcolor}
\usepackage[bookmarksnumbered=true, pagebackref=false]{hyperref}
\definecolor{seagreen}{rgb}{0.1,0.92,0.21}
\definecolor{darkgreen}{rgb}{0.21,0.82,0.11}
\definecolor{deepblue}{rgb}{0.12,0.25,0.62}
\definecolor{darkred}{rgb}{0.92,0,0.12}
\definecolor{lightblue}{RGB}{229,248,255}
\definecolor{lightgray}{RGB}{242,242,242}
\definecolor{lightgray1}{RGB}{180,180,180}

\hypersetup{colorlinks,
	citecolor=[rgb]{0,0.3,0.7},
	filecolor=[rgb]{0,0.1,0.4},
	linkcolor=[rgb]{0.7,0.1,0.2},
	urlcolor=[rgb]{0.9,0.1,0.2}
}

\usepackage{paralist}

\makeatletter
\def\whline#1{%
	\noalign{\ifnum0=`}\fi\hrule \@height #1 \futurelet
	\reserved@a\@xhline}

\begin{document}

\title{Self-Supervised Diffusion Models for Handwritten Signature Verification via Ruin Reconstruction}

\author{Peirong Zhang,
	Zhenhua Yang,
	Jiaxin Zhang,
	Moises Diaz,
	Miguel A. Ferrer,\\
	Umapada Pal,~\IEEEmembership{Senior Member,~IEEE},
	Wei Zhou,~\IEEEmembership{Senior Member,~IEEE},
	Lianwen Jin$^*$~\IEEEmembership{Member,~IEEE}
% <-this % stops a space
\thanks{Peirong Zhang, Zhenhua Yang, Jiaxin Zhang, and Lianwen Jin are with the School of Electronic and Information Engineering, South China University of Technology, Guangzhou, China. (E-mail: eeprzhang@mail.scut.edu.cn; eelwjin@scut.edu.cn)}
\thanks{Moises Diaz and Miguel A. Ferrer are with the Instituto Universitario para el Desarrollo Tecnológico y la Innovación en Comunicaciones, Universidad de Las Palmas de Gran Canaria, Las Palmas, Spain.}
\thanks{Umapada Pal is with the Computer Vision and Pattern Recognition Unit, Indian Statistical Institute, Kolkata 700108, India.}
\thanks{Wei Zhou is with the School of Computer Science and Informatics, Cardiff University, Cardiff, UK.}
\thanks{$^*$ Corresponding author.}
}

\markboth{Under Review}%
{Shell \MakeLowercase{\textit{et al.}}: A Sample Article Using IEEEtran.cls for IEEE Journals}


\maketitle

\begin{abstract}
Handwritten signature verification (HSV) is a critical biometric technique in legal and forensic authentication, whose efficacy hinges on the discriminability of signature representations. Current HSV methods predominantly adopt supervised learning to enhance representation quality, leaving self-supervised learning largely unexplored. We propose \emph{SigDiffuser}, a diffusion-based self-supervised framework for signature representation and verification. At its core, \emph{SigDiffuser} introduces a novel pretext task termed \emph{Reconstruction from Ruins (RFR)}, which deliberately ``ruins'' signatures through severe crumpling, obliterating their structural integrity and writing styles. A visual encoder is employed to capture crucial handwriting patterns from intact signature regions, guiding the denoiser to reconstruct the ``ruined'' portions and learning meaningful representations. Additionally, we introduce a \emph{pyramidal feature aggregation module (PFAM)} to bridge the visual encoder and denoiser, enhancing representation quality via hierarchical aggregation and modulation. Extensive experiments on multilingual signature datasets demonstrate that \emph{SigDiffuser} significantly outperforms both state-of-the-art SSL methods and HSV models. We further show that increasing the crumpling severity on signatures effectively improves representation quality. Notably, our work demonstrates that extreme destruction, coupled with self-supervised diffusion models, can yield superior signature representation compared to traditional mild deformations, paving new avenues for HSV and other biometric domains. Code will be publicly available.
\end{abstract}


\section{Introduction}
\IEEEPARstart{H}{andwritten} signature verification (HSV) has been a longstanding and critical biometric means for distinguishing genuine signatures from forgeries \cite{diaz2019aperspective}. Depending on signature's properties, this technique can be broadly categorized into online and offline forms. Online HSV \cite{msds2022zhang,spectrum2025zhang} processes dynamically captured digital signatures, while the offline counterpart \cite{granulometric1997sabourin,transosv2024li} analyzes static signatures scanned or photographed from paper. Despite the growing prevalence of electronic signing devices, the traditional use of offline signatures on physical documents is still prevalent across various domains such as financial transactions, legal proceedings, and government administration. Consequently, offline HSV remains crucial for ensuring the security and trustworthiness of these applications. In this paper, we focus on offline HSV.

Over the past few decades, the growing demand for real-world HSV has spurred substantial efforts to enhance the reliability of HSV systems. Mainstream methods involve feature representation extraction followed by similarity measurement, in which learning high-quality representations is the key to building a reliable HSV system. Supervised learning techniques have been intensively studied to enhance feature extraction, such as local pattern learning with CNN \cite{idn2019wei} and global contextual modeling using Transformer \cite{transosv2024li}. However, the surge of self-supervised representation learning in other biometric domains \cite{gaitssb2023fan,camd2026} presents a promising avenue for HSV, particularly given the limited availability of HSV data due to the challenge of collecting skilled forgeries. Despite its potential, self-supervised learning remains largely shadowed in the HSV domain, with only a few studies delving into its effectiveness thus far \cite{surds2022icpr,swis2022icip}.

\begin{figure*}[t]
	\centering
	\includegraphics[width=\textwidth]{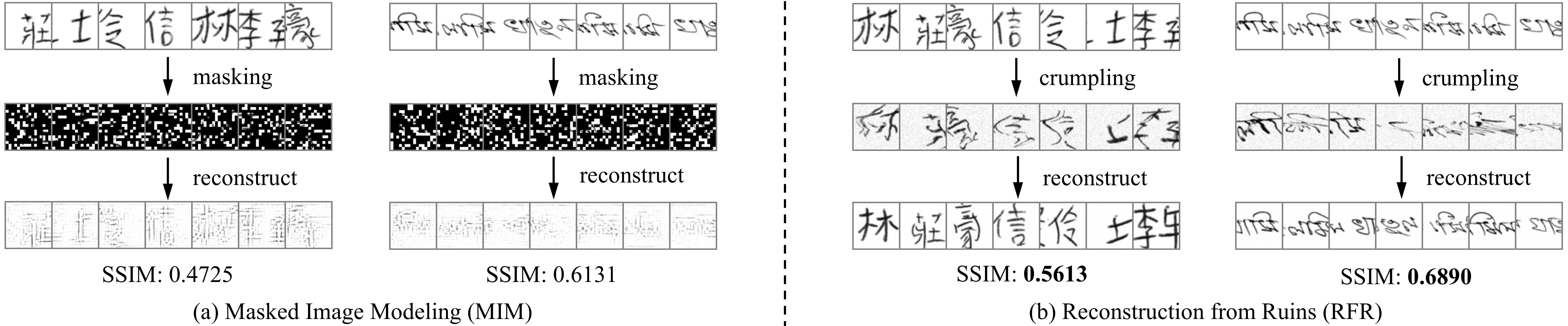}
	\caption{Qualitative and quantitative (Structural Similarity Index (SSIM)) comparisons between MIM and our proposed RFR. (a) MIM exhibits blurred, oversimplified structures that fail to recover complex strokes, resulting in suboptimal representations. (b) In contrast, RFR faithfully reconstructs stroke details and character structures despite severe crumpling, achieving higher SSIM metrics. Although not exact replicas, the fidelity suffices for capturing effective signature representation.}
\label{Fig::comp}
\end{figure*}

Generative representation learning is currently one of the dominant self-supervised learning paradigms in computer vision, as opposed to contrastive learning \cite{sslsurvey2023liu}. Following the principle of ``comprehension by generation'', it typically reconstructs corrupted (masked/augmented) images to extract transferable feature representations for downstream tasks. Masked image modeling (MIM), standing as a prominent exemplar within this paradigm, has achieved remarkable success across myriad vision tasks \cite{mae2022he,simim2022xie}. However, as exemplified in Fig.~\ref{Fig::comp} (a), MIM struggles to accurately restore fine-grained textures and strokes within signatures, yielding only blurred edges and oversimplified structures. This suggests MIM is a suboptimal pretext task for learning effective signature representation.

Beyond MIM, diffusion-based approaches \cite{diffae2022preechakul,ddae2023xiang,soda2024hudson,ldae2025chen} have emerged as a potent force for generative representation learning. Diffusion models progressively add noise to corrupt the clean image into pure Gaussian noise in a forward process, then generate targets via backward denoising. Recent studies \cite{soda2024hudson,diffae2022preechakul,pdae2022zhang} exploit visual encoders alongside the denoiser to extract latent codes from auxiliary images, aiming to assist synthesis and facilitate representation learning. Intuitively, akin to how one discerns the genuineness and authorship of a partially crumpled signature from its visible regions, the corruption process of diffusion models parallels the physical crumpling of a signature, while the visual encoders' latent encoding closely resembles human visual perception of intact regions. This congruence offers a compelling routine for signature representation learning, in which the diffusion model can be tasked with reconstructing the ``crumpled'' part through iterative denoising, guided by the visual features of ``uncrumpled'' details extracted by an additional visual encoder. This approach allows the visual encoder to perceive individualized signature patterns and encode them into latent features, achieving representation learning. Furthermore, the diffusion model's accurate reconstruction of signature details facilitates global visual perception, thereby refining representation quality.

Motivated by this insight, we propose SigDiffuser, a novel diffusion framework that leverages a visual encoder to condition a diffusion denoiser for signature representation learning. SigDiffuser introduces two key innovations. First, we propose \textbf{Reconstruction from Ruins} (RFR; Fig.~\ref{Fig::comp} (b)), a novel pretext task that reconstructs a severely corrupted signature region using a spatially distinct clean portion. Unlike existing pretext tasks that employ mild deformation (\emph{e.g.}, color jittering, image rotation)~\cite{siman2022luo,diffae2022preechakul,ddae2023xiang,soda2024hudson}, RFR inflicts an unprecedented level of corruption, \emph{i.e.}, crumpling, upon signatures, obliterating their text structures and writing styles into a ``ruined'' state. By compelling SigDiffuser to restore these ``ruins'', RFR incentivizes the visual encoder to perceive sufficient writing traits of the clean portion to aid in reconstruction, thereby enhancing latent representation quality. Second, existing methods \cite{diffae2022preechakul,soda2024hudson} bridge the encoder and denoiser with a single latent vector, limiting representation capacity. We address this by introducing a \textbf{pyramidal feature aggregation module} (PFAM) that aggregates multi-scale encoder features into hierarchical latent representations. These features are injected into corresponding denoiser layers for fine-grained modulation, empowering the encoder to learn more informative representations.

We adopt HSV as the downstream task and evaluate SigDiffuser on multiple HSV datasets spanning different languages, including BHSig260-B (Bengali) \cite{bhsig2016pal}, BHSig260-H (Hindi) \cite{bhsig2016pal}, HanSig (Chinese) \cite{hansig2023huang}, UTSig (Persian) \cite{utsig2017soleimani}, and CEDAR (English) \cite{cedar2004kalera}. Extensive experiments demonstrate that SigDiffuser significantly outperforms not only existing self-supervised representation learning methods but also state-of-the-art (SOTA) HSV models. We further observe that increasing the deformation degree enhances representation quality, substantiating the effectiveness of applying severe destruction within the RFR task. 

Our main contributions are summarized as follows:
\begin{itemize}
	\item We propose SigDiffuser, a pioneering framework that integrates diffusion models with self-supervised learning for signature representation. It leverages a visual encoder to capture unique writing patterns from intact signature regions, enabling effective reconstruction of severely crumpled areas and learning informative representations.
	\item We propose a novel Reconstruction from Ruins (RFR) pretext task tailored for signature representation learning. RFR applies severe signature crumpling rather than mild corruptions, effectively incentivizing representation learning by reconstructing severely crumpled signatures.
	\item We introduce a pyramidal feature aggregation module (PFAM) to strengthen representation learning. Through multi-scale feature aggregation and hierarchical modulation, PFAM enables SigDiffuser to extract richer latent representations than single latent embedding approaches.
	\item Experiments demonstrate SigDiffuser's superior performance over both existing representation learning methods and SOTA HSV models. Furthermore, we reveal that increasing the crumpling severity improves representation quality, substantiating the effectiveness of the RFR task.
\end{itemize}

\section{Related Work}
\label{sec::related}
\subsection{Generative Representation Learning}
\subsubsection{Non-Diffusion-Based Methods}
Early generative attempts mainly exploit AutoEncoders and Generative Adversarial Networks (GAN) \cite{dcgan2015} as latent feature learners. Inspired by masked language modeling, the focus transitions to masked image modeling (MIM). Context Encoders \cite{contextinpaint2016pathak} pioneers MIM via image inpainting while Vision Transformer (ViT) \cite{vit2021dosovitskiy} advances to predict the mean color of masked patches. SimMIM \cite{simim2022xie} and MAE \cite{mae2022he} revisit MIM by processing only unmasked patches or all patches, respectively. SparK \cite{spark2023tian} adopts CNN for MIM, in contrast to ViT. Besides, iGPT \cite{igpt2020chen} introduces autoregressive prediction to sequentially predict unknown pixels. SimAN \cite{siman2022luo} restores augmented scene text by aligning styles from its neighboring patches.

\subsubsection{Diffusion-Based Methods}
Beyond their renowned generation fidelity, diffusion models have emerged as powerful representation learners \cite{diffrepsurvey2024}. DiffAE \cite{diffae2022preechakul} conditions the denoiser on a visual encoder to capture semantic traits. DDAE \cite{ddae2023xiang} unifies self-supervised generative and discriminative pretraining via a single denoiser. SODA \cite{soda2024hudson} condenses latent representations using an information bottleneck. \emph{l}-DAE \cite{ldae2025chen} employs a PCA for latent space encoding. DBAE \cite{dbae2025kim} bridges latent features with the generation endpoint to retain complete source information. Unlike prior methods that use single representations to bridge the encoder and denoiser (\emph{e.g.}, DiffAE, SODA, and DBAE), we modulate the denoiser using multi-scale representations aggregated via a feature pyramid to enhance feature extraction. Additionally, we propose a novel RFR pretext task, conditioning the denoiser on severely corrupted signature parts, enforcing the encoder to capture sufficient writing features for restoration.

\subsection{Offline Handwritten Signature Verification}
Offline HSV methods typically consist of two stages: feature extraction and similarity measurement. Based on the feature extraction technique, existing methods can be grouped into hand-crafted feature-based and neural network-based methods.

\subsubsection{Hand-Crafted Feature-Based Method}
Conventional HSV relies heavily on hand-crafted feature descriptors, generally categorized as shape-based and texture-based. Shape-based features capture global appearance and geometric structure \cite{exteriorcontour2005chen,beziercurve2010bertolini,dccm2016hamadene}. Texture-based features capture local texture patterns and micro-structures inside signatures \cite{texturegrey2011vargas,gray2012ferrer,lbp2013hu}. Kumar et al. \cite{surroundedness2012kumar} model both texture and shape with surroundedness features. Other works also explore using KAZE-based Fisher Vectors \cite{kaze2018okawa} and learning SPD-manifold distances from covariance matrices \cite{spd2024zois}. For similarity measurement, Support Vector Machines \cite{texturegrey2011vargas,gray2012ferrer,dccm2016hamadene}, ensemble methods \cite{beziercurve2010bertolini,lbp2013hu} and manifold-based distance measures \cite{spd2024zois} are explored.

\begin{figure*}[t]
	\centering
	\includegraphics[width=\textwidth]{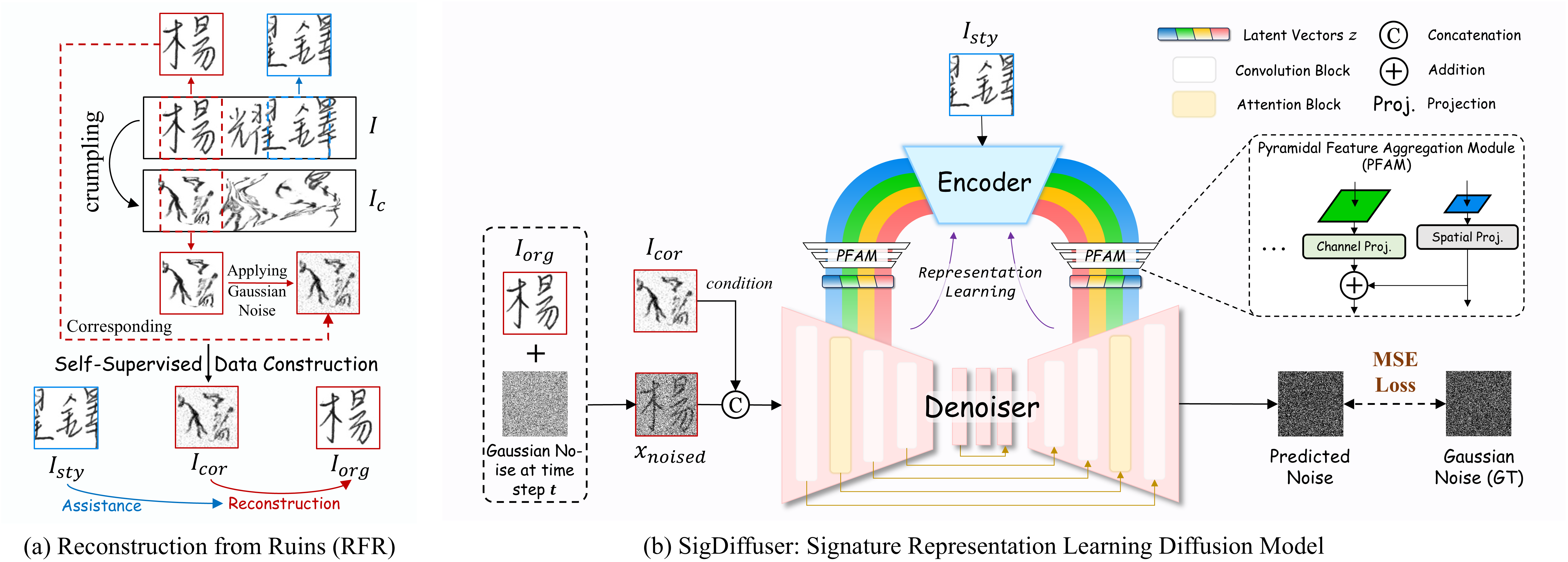}
	\caption{Overall framework of SigDiffuser. (a) Signature data construction for RFR. We apply crumpling on the signature image $I$ to obtain $I_c$ and segment square patches in different positions to acquire $I_{org}$, $I_{sty}$, and $I_{cor}$ for RFR pretraining. (b) Detailed model architecture and self-supervised pretraining process. The visual encoder encodes the intact region $I_{sty}$ into latent representations. These features are hierarchically modulated to the denoiser to assist in reconstructing the severely crumpled portion $I_{cor}$, thereby enforcing the visual encoder to learn meaningful representations.}
	\label{Fig::arch}
\end{figure*}

\subsubsection{Neural Network-Based Method}
Currently, deep neural networks dominate HSV due to their powerful representation capabilities. Early methods used CNNs as simple feature extractors \cite{signetf2017hafemen}, while recent works combine Siamese architectures with CNNs for local feature learning \cite{signet2017dey,idn2019wei,deephsv2019li,msdn2021liu,avn2022li,cwa2024ling}. For instance, MSDN \cite{msdn2021liu} integrates region-based metric learning with a Siamese DenseNet. Subsequently, prompted by its effectiveness, ViT has been integrated into Siamese networks for global feature learning, utilizing mutual attention \cite{transosv2024li} or hybrid CNN-ViT architectures \cite{detailsemnet2024eccv,htcsignet2025zheng}. Two-channel models that process stacked signatures represent an alternative routine \cite{deephsv2019li,2c2s2023ren}.  Beyond architecture, representations are further enriched by using text recognition for multi-task learning \cite{ocr4hsv2024lin} or fusion with handcrafted features \cite{dcccae2025zhao}. For similarity measurement, the paradigm evolves from decoupled verifiers (distance-based \cite{signet2017dey} or SVM-based \cite{signetf2017hafemen,pdsn2018lai}) to end-to-end methods (thresholded sigmoid/softmax functions \cite{msdn2021liu,transosv2024li,cwa2024ling}).

However, existing methods overwhelmingly concentrate on supervised learning, neglecting self-supervised learning's potential in this field. While a few works have explored self-supervision \cite{surds2022icpr,swis2022icip}, such as reconstruction with attention guidance \cite{surds2022icpr} and feature decorrelation \cite{swis2022icip}, they yield inferior HSV performance. In this paper, we propose the first diffusion-based signature representation learning scheme, establishing a new paradigm for HSV through the integration of diffusion models and self-supervised learning.

\section{Methodology}
Fig.~\ref{Fig::arch} illustrates SigDiffuser's overall framework. To extract deep representations, we introduce the Reconstruction from Ruins (RFR) pretext task that compels the model to reconstruct a severely crumpled signature portion using an intact reference. We harness the diffusion process to fulfill the reconstruction, with a pyramidal feature aggregation module (PFAM) to boost representation quality. The pretrained encoder is transferred to downstream HSV tasks.

\subsection{Pretext Task: Reconstruction from Ruins}
\label{sec::pretext}
The aim of RFR lies in reconstructing an utterly ruined signature through the assistance of an intact portion, thus enabling signature representation learning. To this end, we construct pairwise ruined-intact data. Given a clean and flat signature image $I \in \mathbb{R}^{1 \times H \times W}$ ($H$ is height and $W$ is width), we generate a displacement field $D$ with a size of $2\times H \times W$ (2 is the channel number) to emulate crumpling. This field assigns a 2D vector $(dx, dy) = (dx_i,dy_i), i \in \{0,...,H \times W - 1\}$ to each pixel $i$, dictating its horizontal and vertical distortion. $(dx,dy)$ is further processed and applied to clean signatures to create the crumpled outputs. This process is formulated as:

\begin{equation}
\label{eq::alpha}
\begin{gathered}
	D = (dx,dy) = (\mathcal{R}(H,W) \times 2 - 1) \times \alpha,
\end{gathered}
\end{equation}%\vspace{-10pt}
\begin{equation}
\label{eq::theta}
\begin{gathered}
	\mathcal{G}(i,j,\sigma) = \frac{1}{2\pi\sigma^2}e^{-\frac{i^2+j^2}{2\sigma^2}},\\
	\begin{aligned}
		dx_s&=(dx*\mathcal{G})(i,j)\\
		&=\sum\nolimits^{K-1}_{i=0}\sum\nolimits^{K-1}_{j=0} dx(x-i,y-j)\cdot\mathcal{G}(i,j,\sigma),
	\end{aligned}\\
	\begin{aligned}
		dy_s&=(dy*\mathcal{G})(i,j)\\
		&=\sum\nolimits^{K-1}_{i=0}\sum\nolimits^{K-1}_{j=0} dy(x-i,y-j)\cdot\mathcal{G}(i,j,\sigma),
	\end{aligned}
\end{gathered}
\end{equation}
where $\mathcal{G}$ specifies a Gaussian kernel (size $K$, standard deviation $\sigma$), $\mathcal{R}$ denotes a random noise generation function, and $\mathcal{R}(H,W)$ generates a random displacement field with a size of $H\times W$. $\alpha$ acts as a scale factor. By prescribing extremely large values for $\alpha$, $\sigma$, and $K$, we induce a highly intense elastic distortion field $dx$ and $dy$ along the $x$ and $y$ axes as per Eq.~\ref{eq::alpha}. Thereafter, we apply large-kernel Gaussian smoothing to ensure continuity as in Eq.~\ref{eq::theta}. The smoothed distortion fields then undergo random shifts and scaling, with pixel adjustments based on their distance from the image center, yielding modified displacement fields $dx_s'$ and $dy_s'$. Ultimately, the distorted displacement fields $dx_s'$ and $dy_s'$ are added to the original pixel coordinates to obtain the ``crumpled'' coordinates, and we remap $I$ according to these coordinates to create a severely crumpled signature $I_c$:
\begin{equation}
\begin{gathered}
	D=[dx_s',dy_s'],\\
	I_c = I'(x,y) = I(x+D(0,x,y), y+D(1,x,y)).
\end{gathered}
\end{equation}

Following this, we apply Gaussian noise to the crumpled signatures. We then crop a square patch $I_{org}$ of side length $l$ (the shorter side of $I$) at a random position of $I$, and crop a corresponding crumpled patch $I_{cor}$ from $I_c$ at the same position. An equal-sized patch $I_{sty}$ is cropped from $I$ elsewhere to serve as the style reference. This process is illustrated in Fig.~\ref{Fig::arch} (a). Mirroring the human cognitive process of inferring holistic writing styles from visible parts, RFR requires the SigDiffuser model to restore the crumpled patch $I_{cor}$ to its clean version $I_{org}$ under the guidance of $I_{sty}$. We use a visual encoder $\mathcal{E}$ (detailed in the next subsection) to extract writing style representations $z_{sty}=\mathcal{E}(I_{sty})$ of the same signature from $I_{sty}$ to aid in reconstruction.

\textbf{Technical foundation of RFR: shortcut-resistant information bottleneck}. Technically, RFR induces a shortcut-resistant information bottleneck by suppressing the direct information path from $I_{cor}$ to $I_{org}$, thereby making reconstruction depend more on the style representation $z_{sty}$. Let $\mathcal{I}(\cdot;\cdot)$ denote mutual information. The reconstruction-relevant information can be decomposed as:
\begin{equation}
	\mathcal{I}(I_{org}; z_{sty}, I_{cor})
	= \mathcal{I}(I_{org}; I_{cor})
	+ \mathcal{I}(I_{org}; z_{sty} \mid I_{cor}).
	\label{eq::mi_chain}
\end{equation}

In self-supervised tasks with mild image transformations \cite{imagerotate2018spyros,siman2022luo} or simple noise addition \cite{diffae2022preechakul,ddae2023xiang,soda2024hudson}, the corrupted input preserves strong local structural correspondence with the target. Similarly, conventional reconstruction paradigms like Masked Image Modeling (MIM) \cite{mae2022he,simim2022xie} lack an explicit, spatially independent style reference, relying heavily on local spatial cues from unmasked patches. In these cases, the first term $\mathcal{I}(I_{org}; I_{cor})$ remains dominant, allowing the denoiser/generator to exploit low-level pixel or geometry shortcuts and reducing the learning pressure on the encoder. In contrast, the severe crumpling introduced in RFR destroys local stroke continuity and spatial layouts, driving the direct target-corrupted mutual information toward a small residual, \emph{i.e.}, $\mathcal{I}(I_{org}; I_{cor})\rightarrow\epsilon$, forcing the reconstruction predominantly rely on $\mathcal{I}(I_{org}; z_{sty} \mid I_{cor})$. 

To ensure $z_{sty}$ provides meaningful guidance for this conditional dependence, the source patch $I_{sty}$ is intentionally cropped from a spatially distinct position from $I_{org}$. This prevents the visual encoder from trivially copying pixels, compelling it to encode intrinsic writing traits into $z_{sty}$, such as stroke dynamics (stroke connections and turns), slant tendency, and pressure variations, that transfer across spatial locations. In this context, $z_{sty}$ supplies rich style cues to maximize the conditional mutual information $\mathcal{I}(I_{org}; z_{sty} \mid I_{cor})$, effectively preventing superficial pixel shortcuts during reconstruction and improving the encoder's representation quality.

\subsection{SigDiffuser}
\label{sec::sigdiffuser}
We present the architecture of the SigDiffuser model in Fig.~\ref{Fig::arch} (b), consisting of an encoder $\mathcal{E}$, a pyramidal feature aggregation module (PFAM) $\mathcal{P}$, and a denoiser $\mathcal{D}$.

\textbf{Encoder.} The encoder $\mathcal{E}$ is a ResNet-18 \cite{resnet2016he} without the second and last downsampling operations. It encodes the style signature patch $I_{sty}$ to derive hierarchical feature maps:
\begin{equation}
\label{eq::encoder}
\{f_i\} = \{f_1,f_2,...,f_n\} = \mathcal{E}(I_{sty}), i \in \{1,...,n\},
\end{equation}
where $n$ denotes the number of hierarchies in $\mathcal{E}$. These feature maps are then fed into the PFAM for hierarchical aggregation.

\textbf{Pyramidal feature aggregation module.} Existing models like DiffAE \cite{diffae2022preechakul} and SODA \cite{soda2024hudson} input only a single latent representation (typically from the last layer) to the denoiser, leading to limited representational capacity. To address this issue, we introduce the PFAM ($\mathcal{P}$; see Fig.~\ref{Fig::arch} (b)) to extract richer representations by hierarchically aggregating multi-scale features from different layers of the visual encoder. PFAM is simplified from the feature pyramid network \cite{fpn2017lin} by removing the final $3\times3$ convolution on the merged feature map. It takes the feature maps $\{f_i\}$ outputted by the encoder $\mathcal{E}$, deriving a set of compact latent representations $z=\{z_i\}, i \in \{1,...,n\}$:
\begin{equation}
	\label{eq::pfam}
	z_i = \begin{cases}
		\mathcal{P}_i(f_i)= \psi_i(f_i), \text{if}~i=1,\\
		\mathcal{P}_i(f_{i-1},f_i)= \psi_i(f_{i-1})\oplus\phi_i(f_i), \text{otherwise},
	\end{cases}
\end{equation}
where $\mathcal{P}_i$ is the $i^{th}$ layer of the PFAM, $\psi_i$ is a $1\times1$ convolution layer for channel projection, $\phi_i$ is an upsample operation for spatial dimension matching through bilinear interpolation, and $\oplus$ is element-wise addition. Through this hierarchical aggregation process (Eq.~\ref{eq::pfam}), the $i^{th}$ layer of PFAM amalgamates feature maps $f_{i-1}$ and $f_i$ to encode multi-scale writing traits, producing a more informative latent representation $z_i$ at the current hierarchy. These hierarchical embeddings $z=\{z_i\}$ are subsequently injected into different layers of the denoiser $\mathcal{D}$ for layer-wise modulation. $n$ is set to 4.

\textbf{Denoiser.} We build the denoiser $\mathcal{D}$ employing a conditional Denoising Diffusion Implicit Model (DDIM) \cite{ddim2021song}, which is conditioned on the hierarchical latent representations $z=\{z_i\},i \in \{1,...,n\}$. It incorporates a forward process that progressively corrupts the target distribution $\boldsymbol{x}_0 \sim q(\boldsymbol{x}_0)$ into pure Gaussian noise $\boldsymbol{x}_T \sim \mathcal{N}(0,\boldsymbol{I})$, complemented by a backward process that reconstructs the target by $T$ steps iterative denoising from the conditional input $\boldsymbol{x}_T'=(\boldsymbol{x}_T,z)$.

The forward process is a Markov chain that incrementally adds noise to the target, as formulated by:
\begin{equation}
\begin{gathered}
	\boldsymbol{x}_t = \sqrt{\bar{\alpha_t}}\boldsymbol{x}_0 + \sqrt{1-\bar{\alpha_t}}\boldsymbol{\epsilon};~
	\bar{\alpha}_t=\prod\nolimits^t_{s=1}(1 - \beta_s),
\end{gathered}
\end{equation}
where $t\sim[0,T]$, $T$ is the maximum sampling timestep. $\beta\sim(0,1)$ is a fixed hyperparameter that indicates the noise levels. The backward process takes $\boldsymbol{x}_T'=(\boldsymbol{x}_T,z)$ as input and deterministically approximates the reverse mapping:
\begin{equation}
\label{eq::noise}
\begin{aligned}
	\boldsymbol{x}_{t-1}=&\sqrt{\bar{\alpha}_{t-1}}\frac{\boldsymbol{x}_t-\sqrt{1-\bar{\alpha}_t}\boldsymbol{\epsilon}_\theta^t(\boldsymbol{x}_t,t,z)}{\sqrt{\bar{\alpha}_t}} + \\
	&\sqrt{1 - \bar{\alpha}_{t-1}}\boldsymbol{\epsilon}_\theta^t(\boldsymbol{x}_t,t,z).
\end{aligned}
\end{equation}

The noise term $\boldsymbol{\epsilon}_\theta^t(x_t,t,z)$ in Eq.~\ref{eq::noise} is predicted by the denoiser under the supervision of the MSE loss $||\boldsymbol{\epsilon}_\theta^t(x_t,t,z)-\boldsymbol{\epsilon}||$. As illustrated in Fig.~\ref{Fig::arch} (b), the denoiser adopts a four-layer, symmetric UNet architecture \cite{unet2015ronneberger}. Since we compute four latent representations $\{z_i\}, i \in \{1,...,4\}$, we symmetrically add $z_i$ into the $i^{th}$ UNet layer through addition, achieving hierarchical modulation. From a global perspective, the denoiser is primarily responsible for image reconstruction, thus liberating the encoder to focus on latent encoding. Given the demanding reconstruction task (\emph{i.e.}, RFR), the encoder is driven to leverage any available writing features to assist the denoiser, which in turn incentivizes itself to extract more informative writing patterns, resulting in enhanced latent representations. 

Notably, closer to our work are SODA \cite{soda2024hudson} and DiffAE \cite{diffae2022preechakul}, yet SigDiffuser differs in two main aspects. (1) For the pretext task, SigDiffuser reconstructs severely crumpled signatures (the proposed RFR). However, SODA synthesizes a novel view of the image, while DiffAE performs reconstruction with the same image as semantic guidance. (2) Regarding architecture, SigDiffuser introduces a multi-scale, hierarchical feature modulation architecture for enriched latent features, whereas both SODA and DiffAE employ a single-representation approach. These approaches together result in better self-supervised learning performance on HSV as validated in Table~\ref{Table::finetune}.
%distilling the essence of the input signature into enriched latent representations.

\begin{figure}[t]
	\centering
	\includegraphics[width=\linewidth]{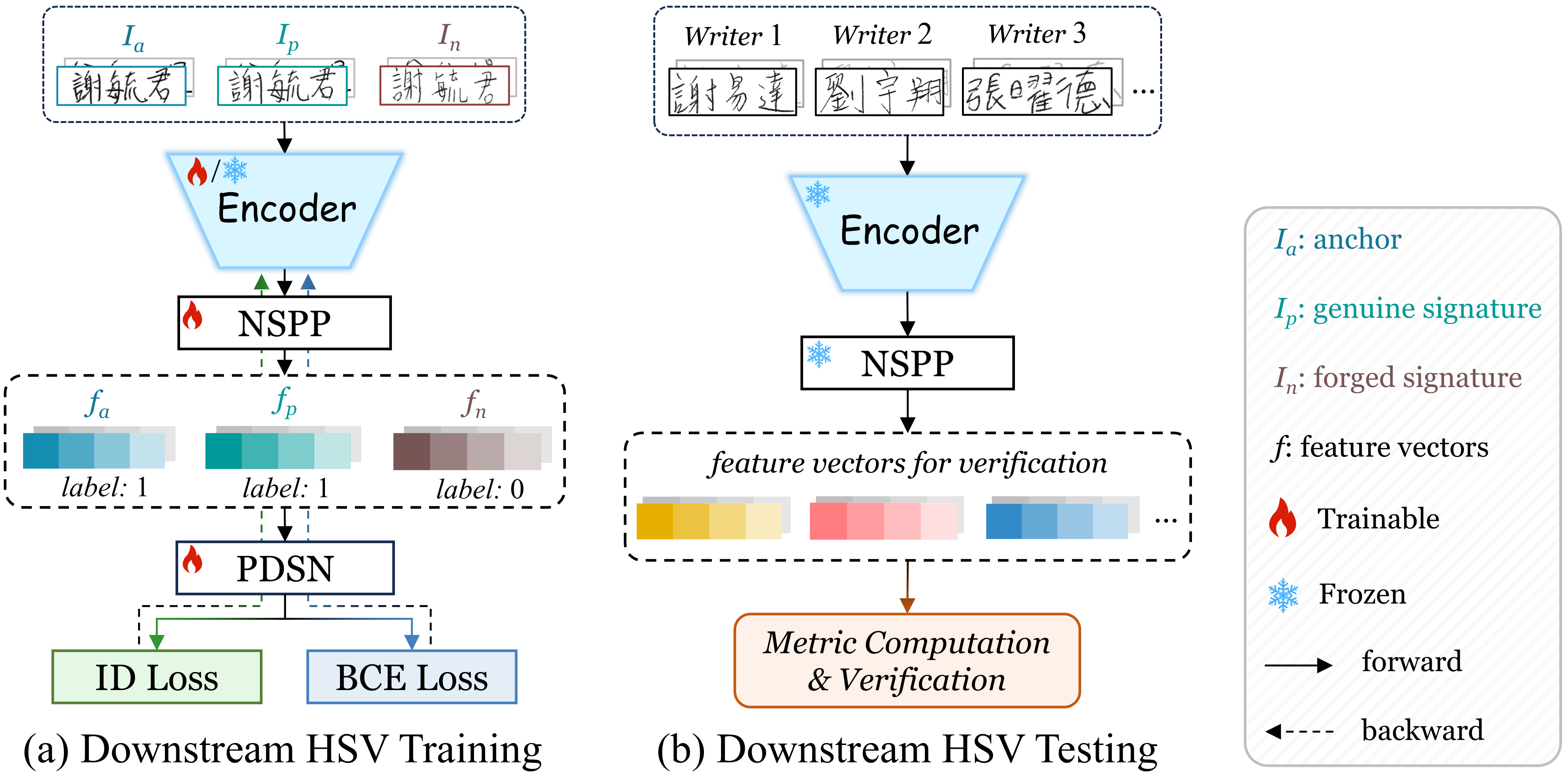}
	\caption{Downstream model architecture for HSV. (a) Training. The encoder pretrained on the RFR pretext task is transferred to the downstream model for HSV fine-tuning. (b) Testing. All signatures are input to the model to extract corresponding feature vectors, which are then used for verification.}
	\label{Fig::downstream}
\end{figure}

\begin{table*}[t]
	\renewcommand{\arraystretch}{1}
	\caption{Dataset details and data splitting. {\#Writer\tiny{Training}} and {\#Writer\tiny{Testing}} respectively denote the number of writers for training and testing. Sig. denotes signature. /writer denotes per writer.}
	\label{Table::dataset}
	\centering
	\resizebox{\textwidth}{!}{
		\begin{tabular}{l c c c c c c c c c c}
			\whline{1.05pt}
			{Dataset} & {Language} & {\#Writer} & {\#Writer\tiny{Training}} & {\#Writer\tiny{Testing}} & {\#Genuine Sig./writer} & {\#Forged Sig./writer} & {\#Split} & {\#Training} & {\#Testing} & {\#Total}\\
			\hline
			BHSig260-B \cite{bhsig2016pal} & Bengali & 100 & 50 & 50 & 24 & 30 & 5 & 2,700 & 2,700 & 5,400\\
%			\hline
			BHSig260-H \cite{bhsig2016pal} & Hindi & 160 & 100 & 60 & 24 & 30 & 5 & 5,400 & 3,240 & 8,640\\
%			\hline
			HanSig \cite{hansig2023huang} & Chinese & 238 & 213 & 25 & 20 & 20 & 10 & 31,800 & 3,600 & 35,400\\
%			\hline
			UTSig \cite{utsig2017soleimani} & Persian & 115 & 60 & 55 & 27 & 45 & 5 & 4,320 & 3,960 & 8,280\\
%			\hline
			CEDAR \cite{cedar2004kalera} & English & 55 & 50 & 5 & 24 & 24 & 11 & 2,400 & 240 & 2,640\\
			\whline{1.05pt}
	\end{tabular}}
\end{table*}

\subsection{Downstream HSV Model}
\label{sec::downstream}
After the self-supervised pretraining with the RFR pretext task, we fine-tune a specific model for the downstream HSV task, as illustrated in Fig.~\ref{Fig::downstream}. This model's fine-tuning is independent of SigDiffuser's pretraining (Sec.~\ref{sec::sigdiffuser}), and all testing results (Sec.~\ref{sec::exp}) are obtained using this model. Model architecture, training, and testing details are elaborated below.

\textbf{Model architecture}. As shown in Fig.~\ref{Fig::downstream} (a), the encoder of SigDiffuser pretrained via the RFR pretext task transfers directly as the HSV model encoder. Inspired by Spatial Pyramid Pooling (SPP) \cite{spp2015he}, we design an improved Normalized SPP (NSPP) module, which is placed after the encoder to reduce the spatial dimension of the last feature map into feature vectors. Given the final feature map $x \in \mathbb{R}^{512\times h\times w}$, where 512 is the channel number, $h$ and $w$ respectively denote the height and width, it is first compressed by average pooling to derive local and global features $x_l$ and $x_g$:
\begin{equation}
	\begin{gathered}
		x_g = \text{AdaAvgPool}_{2\times2}(x),~
		x_l = \text{AdaAvgPool}_{1\times1}(x).
	\end{gathered}
\end{equation}
To ensure scale invariance and enhance discriminability, both features are L2-normalized to obtain $\widetilde{x}_g$ and $\widetilde{x}_l$. They are then transformed into 1D vectors using convolutional and fused:
\begin{equation}
	\begin{gathered}
		\hat{x}_g = \text{conv}_{2\times2}(\widetilde{x}_g),~\hat{x}_l = \text{conv}_{1\times1}(\widetilde{x}_l),\\
		f = \text{SELU}(\hat{x_g} + \hat{x_l}), f \in \mathbb{R}^{512}.
	\end{gathered}
\end{equation}
The final feature vector $f$ is used for loss computation or identity verification as described below.

\textbf{Downstream model training.} As shown in Fig.~\ref{Fig::downstream}(a), we append a Position-Dependent Siamese Network (PDSN) \cite{pdsn2018lai} after the NSPP module strictly for training. For each writer, we sample triplets $(I_a,I_p,I_n)$ comprising an anchor (genuine), a positive  (genuine), and a negative (skilled forgery) signature. The encoder and NSPP extract corresponding features $(f_a,f_p,f_n)$. PDSN processes a collection of pairs $[(f_a,f_n),(f_p,f_n),(f_a,f_p)]$ to compute a logit triplet, which is supervised by binary cross-entropy loss $\mathcal{L}_{BCE}$ against ground-truth labels $[1,1,0]$. Additionally, the feature vectors pass through a linear layer for writer ID classification, optimized via label-smoothed cross-entropy loss $\mathcal{L}_{ID}$ (label smoothing 0.1). The final optimization objective is $\mathcal{L}=5\mathcal{L}_{BCE} + 0.01\mathcal{L}_{ID}$. We process $B$ triplets from distinct writers per batch, resulting in an effective batch size of 24.

\textbf{Downstream model testing.} During testing, genuine signatures and skilled forgeries are processed to extract feature vectors via the NSPP module (bypassing the PDSN), as shown in Fig.~\ref{Fig::downstream} (b). These features are used to determine the genuineness of each sample and evaluate model performance.

\section{Experiment}
\label{sec::exp}
\subsection{Experimental Protocol}
\subsubsection{Dataset}
\label{sec::dataset}
We evaluate our method on multiple HSV datasets using two data-splitting settings: \emph{(i) Standard split}, following existing training/testing partitions; and \emph{(ii) Split with validation}, which holds out a validation subset from the training data for early stopping and checkpoint selection. As these datasets lack official splits, we mitigate evaluation bias by employing standard K-fold cross-validation (where test folds are mutually exclusive). For datasets indivisible into K folds under prior partitions (\emph{e.g.}, 100/60 for BHSig260-H), we adopt full-coverage random splits, ensuring every writer appears in the test set at least once, with some writers recurring in multiple test splits. These protocols are applied to all datasets, with per-dataset details presented below.

\textbf{BHSig260.} BHSig260 \cite{bhsig2016pal} contains BHSig260-B (5,400 Bengali signatures from 100 writers) and BHSig260-H (8,640 Hindi signatures from 160 writers), with each writer contributing 24 genuine signatures and 30 skilled forgeries. Under setting (i), BHSig260-B is split into 50/50 writers for training/testing, and BHSig260-H is split into 100/60, following \cite{signet2017dey,idn2019wei,avn2022li,transosv2024li}. Under setting (ii), 10 and 20 training writers are further reserved for validation on BHSig260-B and BHSig260-H, respectively, yielding 40/10/50 and 80/20/60 training/validation/testing splits. We perform 5 full-coverage random splits on each of BHSig260-B and BHSig260-H.

%the BHSig260-B and BHSig260-H subsets. BHSig260-B includes 5,400 Bengali signatures from 100 writers and BHSig260-H contains 8,640 Hindi signatures from 160 writers. Each writer has contributed 24 genuine signatures and 30 skilled forgeries. We follow common practices \cite{signet2017dey,idn2019wei,avn2022li,transosv2024li} to split the training and testing data. For BHSig260-B, we randomly allocate the data of 50 writers as the training set and utilize the other 50 writers' data as the testing set. For BHSig260-H, the data from 100 writers is designated as the training set, with the data from the remaining 60 writers used for testing.

\textbf{HanSig.} HanSig \cite{hansig2023huang} is a Chinese signature dataset containing 885 names signed by 238 writers. Each name has 20 genuine signatures in three styles (neat, normal, and stylish) and an equal number of skilled forgeries, totaling 35,400 signatures. Under setting (i), 795 names from 213 writers are used for training and 90 names from 25 writers for testing, following the original protocol. Under setting (ii), 60 training names are reserved for validation, giving a 735/60/90 split. We perform 10 full-coverage random splits.

\textbf{UTSig.} UTSig \cite{utsig2017soleimani} includes 8,280 Persian signatures from 115 individuals, each providing 27 genuine signatures and 45 forgeries (42 skilled forgeries and three opposite-hand written). Under setting (i), 60/55 writers are used for training/testing following the original protocol. Under setting (ii), 10 training writers are reserved for validation, giving a 50/10/55 split. We perform 5 full-coverage random splits.

\textbf{CEDAR.} CEDAR \cite{cedar2004kalera} comprises 2,640 English signatures from 55 writers, with 24 genuine signatures and 24 skilled forgeries per writer. Under setting (i), 50/5 writers are used for training/testing following \cite{msdn2021liu,transosv2024li}. Under setting (ii), 5 training writers are reserved for validation, giving a 45/5/5 split. Unlike the other datasets, which use full-coverage random splits, we perform 11-fold cross-validation on CEDAR.

Data splitting details are summarized in Table~\ref{Table::dataset}. Note that writer sets for training and testing are disjoint, ensuring open-set evaluations. For BHSig260, we pretrain on the union of the BHSig260-B and BHSig260-H training sets, but fine-tune and evaluate on each separately. For other datasets, pretraining and fine-tuning share the corresponding training set. To prevent data leakage, pretraining and fine-tuning are independently rerun per split or fold. We report the mean and standard deviation of the performance as $\text{mean}~(\pm \text{std})$.

%Given the contrasting character structures and idiosyncratic writing styles inherent to different languages, we aim to assess the generalizability of our method on multilingual signatures.

\begin{table*}[t]
	\renewcommand{\arraystretch}{1.0}
	\caption{Linear probing and downstream fine-tuning results on BHSig260-B \cite{bhsig2016pal} and BHSig260-H \cite{bhsig2016pal}. The best and second-best results are marked in \textbf{bold} and with \underline{underlines}, with the same annotations throughout. Superscripts *, $^\dagger$, $^\star$ indicate models utilizing their original backbones: ResNet-50 \cite{resnet2016he}, a modified 10-layer VGG \cite{vgg2015}, and ConvNeXt-Small \cite{convnext2020liu}, respectively.}
	\label{Table::finetune}
	\centering
	\setlength{\tabcolsep}{3pt}
	\resizebox{\textwidth}{!}{
		\begin{tabular}{l c c c c c | c c c c | c c c c | c c c c}
			\toprule
			\multirow{3.6}{*}{{Method}} &
			\multirow{3.6}{*}{{Venue}} &  \multicolumn{8}{c}{Linear Probing} & \multicolumn{8}{|c}{Downstream Fine-tuning}\\
			\cmidrule(r){3-10}\cmidrule(r){11-18}
			~ & ~ & \multicolumn{4}{c}{BHSig260-B} & \multicolumn{4}{c}{BHSig260-H} & \multicolumn{4}{|c}{BHSig260-B} & \multicolumn{4}{c}{BHSig260-H}\\
			\cmidrule(r){3-6}\cmidrule(r){7-10}\cmidrule(r){11-14}\cmidrule(r){15-18}
			~ & ~ & EER $\downarrow$ & FAR $\downarrow$ & FRR $\downarrow$ & Acc $\uparrow$ & EER $\downarrow$ & FAR $\downarrow$ & FRR $\downarrow$ & Acc $\uparrow$ & EER $\downarrow$ & FAR $\downarrow$ & FRR $\downarrow$ & Acc $\uparrow$ & EER $\downarrow$ & FAR $\downarrow$ & FRR $\downarrow$ & Acc $\uparrow$\\
			\hline
			\footnotesize \emph{Contrastive SSRL Methods}\\
			\hline
			SimCLR \cite{simclr2020icml} & ICML'20 & 34.96 ($\pm$0.57) & 34.96 ($\pm$0.57) & 34.96 ($\pm$0.57) & 65.04 ($\pm$0.57) & 35.55 ($\pm$0.54) & 35.56 ($\pm$0.54) & 35.54 ($\pm$0.54) & 64.45 ($\pm$0.54) & 8.28 ($\pm$1.52) & 8.28 ($\pm$1.53) & 8.28 ($\pm$1.52) & 91.72 ($\pm$1.52) & 6.36 ($\pm$1.23) & 6.37 ($\pm$1.22) & 6.36 ($\pm$1.23) & 93.63 ($\pm$1.22)\\
			SimCLR* \cite{simclr2020icml} & ICML'20 & 39.59 ($\pm$0.98) & 39.58 ($\pm$0.99) & 39.59 ($\pm$0.98) & 60.42 ($\pm$0.99) & 38.35 ($\pm$0.68) & 38.35 ($\pm$0.69) & 38.36 ($\pm$0.67) & 61.65 ($\pm$0.68) & 9.67 ($\pm$1.58) & 9.67 ($\pm$1.58) & 9.67 ($\pm$1.58) & 90.33 ($\pm$1.58) & 8.54 ($\pm$1.53) & 8.54 ($\pm$1.54) & 8.54 ($\pm$1.53) & 91.46 ($\pm$1.54)\\
			BYOL \cite{byol2020nips} & NeurIPS'20 & 13.51 ($\pm$1.69) & 13.52 ($\pm$1.69) & 13.51 ($\pm$1.70) & 86.48 ($\pm$1.69) & 27.08 ($\pm$0.90) & 27.08 ($\pm$0.90) & 27.08 ($\pm$0.90) & 72.92 ($\pm$0.90) & 5.86 ($\pm$0.73) & 5.86 ($\pm$0.73) & 5.85 ($\pm$0.73) & 94.14 ($\pm$0.73) & 7.11 ($\pm$1.67) & 7.11 ($\pm$1.67) & 7.11 ($\pm$1.67) & 92.89 ($\pm$1.67)\\
			BYOL* \cite{byol2020nips} & ICML'20 & 23.64 ($\pm$1.69) & 23.64 ($\pm$1.69) & 23.64 ($\pm$1.69) & 76.36 ($\pm$1.69) & 28.37 ($\pm$1.23) & 28.37 ($\pm$1.23) & 28.38 ($\pm$1.23) & 71.63 ($\pm$1.23) & 5.98 ($\pm$1.78) & 5.98 ($\pm$1.78) & 5.98 ($\pm$1.78) & 94.02 ($\pm$1.78) & 8.05 ($\pm$1.98) & 8.05 ($\pm$1.98) & 8.05 ($\pm$1.99) & 91.95 ($\pm$1.98)\\
			SeqCLR \cite{seqclr2021aberdam} & CVPR'21 & 20.25 ($\pm$1.56) & 20.25 ($\pm$1.56) & 20.26 ($\pm$1.56) & 79.75 ($\pm$1.56) & 22.99 ($\pm$0.65) & 22.99 ($\pm$0.64) & 22.99 ($\pm$0.65) & 77.01 ($\pm$0.64) & 10.81 ($\pm$0.67) & 10.82 ($\pm$0.67) & 10.81 ($\pm$0.67) & 89.19 ($\pm$0.67) & 9.51 ($\pm$1.48) & 9.50 ($\pm$1.48) & 9.51 ($\pm$1.48) & 90.50 ($\pm$1.48)\\
			%			\hline
			ChaCo \cite{chaco2022zhang} & ICFHR'22 & 11.10 ($\pm$1.04) & 11.10 ($\pm$1.04) & 11.10 ($\pm$1.04) & 88.90 ($\pm$1.04) & 16.60 ($\pm$0.72) & 16.60 ($\pm$0.71) & 16.60 ($\pm$0.72) & 83.40 ($\pm$0.72) & 6.73 ($\pm$1.83) & 6.73 ($\pm$1.83) & 6.73 ($\pm$1.82) & 93.27 ($\pm$1.83) & 5.35 ($\pm$1.18) & 5.35 ($\pm$1.18) & 5.34 ($\pm$1.18) & 94.65 ($\pm$1.18)\\
			%			\hline
			CMT-Co \cite{cmtco2022zhang} & ACCV'22 & 15.81 ($\pm$1.62) & 15.82 ($\pm$1.62) & 15.81 ($\pm$1.62) & 84.19 ($\pm$1.62) & 19.89 ($\pm$0.99) & 19.90 ($\pm$0.99) & 19.89 ($\pm$0.98) & 80.10 ($\pm$0.99) & 7.35 ($\pm$1.40) & 7.35 ($\pm$1.41) & 7.34 ($\pm$1.40) & 92.65 ($\pm$1.41) & 5.83 ($\pm$1.88) & 5.83 ($\pm$1.88) & 5.83 ($\pm$1.88) & 94.17 ($\pm$1.88)\\
			SWIS \cite{swis2022icip} & ICIP'22 & 13.35 ($\pm$0.65) & 13.35 ($\pm$0.65) & 13.35 ($\pm$0.66) & 86.65 ($\pm$0.65) & 17.65 ($\pm$1.26) & 17.65 ($\pm$1.26) & 17.65 ($\pm$1.26) & 82.35 ($\pm$1.26) & 8.46 ($\pm$1.56) & 8.46 ($\pm$1.56) & 8.45 ($\pm$1.56) & 91.54 ($\pm$1.56) & 9.52 ($\pm$4.50) & 9.52 ($\pm$4.49) & 9.53 ($\pm$4.50) & 90.48 ($\pm$4.49)\\
			\hline
			\footnotesize \emph{Generative SSRL Methods}\\
			\hline
			SURDS \cite{surds2022icpr} & ICPR'22 & 16.17 ($\pm$2.17) & 16.16 ($\pm$2.17) & 16.17 ($\pm$2.17) & 83.84 ($\pm$2.17) & 26.83 ($\pm$0.86) & 26.83 ($\pm$0.86) & 26.83 ($\pm$0.86) & 73.17 ($\pm$0.86) & 7.21 ($\pm$1.05) & 7.21 ($\pm$1.05) & 7.22 ($\pm$1.06) & 92.79 ($\pm$1.05) & 6.20 ($\pm$1.02) & 6.20 ($\pm$1.02) & 6.20 ($\pm$1.02) & 93.80 ($\pm$1.02)\\
			SimAN \cite{siman2022luo} & CVPR'22 & 19.66 ($\pm$1.41) & 19.65 ($\pm$1.41) & 19.66 ($\pm$1.42) & 80.34 ($\pm$1.41) & 30.33 ($\pm$0.72) & 30.33 ($\pm$0.73) & 30.33 ($\pm$0.72) & 69.67 ($\pm$0.72) & 5.93 ($\pm$0.46) & 5.93 ($\pm$0.46) & 5.92 ($\pm$0.46) & 94.07 ($\pm$0.46) & 6.52 ($\pm$1.41) & 6.52 ($\pm$1.41) & 6.52 ($\pm$1.41) & 93.48 ($\pm$1.41)\\
			SimAN$^\dagger$ \cite{siman2022luo} & CVPR'22 & 28.18 ($\pm$1.56) & 28.17 ($\pm$1.58) & 28.15 ($\pm$1.53) & 71.83 ($\pm$1.56) & 36.59 ($\pm$0.49) & 36.62 ($\pm$0.42) & 36.56 ($\pm$0.55) & 63.40 ($\pm$0.46) & 12.14 ($\pm$8.50) & 12.16 ($\pm$8.54) & 12.12 ($\pm$8.44) & 87.86 ($\pm$8.52) & 7.77 ($\pm$1.26) & 7.73 ($\pm$1.30) & 7.83 ($\pm$1.20) & 92.24 ($\pm$1.27)\\
			%			\hline
			MAE \cite{mae2022he} & CVPR'22 & 30.89 ($\pm$2.03) & 30.90 ($\pm$2.03) & 30.88 ($\pm$2.02) & 69.11 ($\pm$2.03) & 41.98 ($\pm$0.74) & 41.98 ($\pm$0.75) & 41.99 ($\pm$0.73) & 58.02 ($\pm$0.74) & 15.24 ($\pm$1.68) & 15.24 ($\pm$1.70) & 15.25 ($\pm$1.66) & 84.76 ($\pm$1.69) & 37.86 ($\pm$1.78) & 38.18 ($\pm$1.47) & 37.57 ($\pm$2.05) & 61.99 ($\pm$1.63)\\
			%			\hline
			DiffAE \cite{diffae2022preechakul} & CVPR'22 & 29.99 ($\pm$3.59) & 29.99 ($\pm$3.59) & 29.99 ($\pm$3.59) & 70.01 ($\pm$3.59) & 29.89 ($\pm$2.59) & 29.89 ($\pm$2.58) & 29.89 ($\pm$2.59) & 70.11 ($\pm$2.58) & 7.40 ($\pm$1.72) & 7.40 ($\pm$1.73) & 7.40 ($\pm$1.72) & 92.60 ($\pm$1.73) & 9.21 ($\pm$1.93) & 9.21 ($\pm$1.93) & 9.21 ($\pm$1.93) & 90.79 ($\pm$1.93)\\
			SparK \cite{spark2023tian} & ICLR'23 & 13.54 ($\pm$1.19) & 13.54 ($\pm$1.19) & 13.54 ($\pm$1.19) & 86.46 ($\pm$1.19) & 20.45 ($\pm$0.85) & 20.45 ($\pm$0.85) & 20.46 ($\pm$0.85) & 79.55 ($\pm$0.85) & 5.42 ($\pm$1.42) & 5.42 ($\pm$1.42) & 5.42 ($\pm$1.41) & 94.58 ($\pm$1.42) & 7.00 ($\pm$1.42) & 7.00 ($\pm$1.42) & 7.00 ($\pm$1.43) & 93.00 ($\pm$1.42)\\
			SparK$^\star$ \cite{spark2023tian} & ICLR'23 & 25.46 ($\pm$1.14) & 25.46 ($\pm$1.13) & 25.47 ($\pm$1.14) & 74.54 ($\pm$1.14) & 35.44 ($\pm$1.27) & 35.44 ($\pm$1.28) & 35.45 ($\pm$1.27) & 64.56 ($\pm$1.28) & 14.16 ($\pm$0.80) & 14.16 ($\pm$0.80) & 14.15 ($\pm$0.79) & 85.84 ($\pm$0.80) & 11.71 ($\pm$1.69) & 11.70 ($\pm$1.69) & 11.71 ($\pm$1.69) & 88.30 ($\pm$1.69)\\
			%			\hline
			DDAE \cite{ddae2023xiang} & ICCV'23 & 10.32 ($\pm$1.23) & 10.32 ($\pm$1.23) & 10.32 ($\pm$1.24) & 89.68 ($\pm$1.23) & 17.11 ($\pm$0.84) & 17.10 ($\pm$0.84) & 17.11 ($\pm$0.84) & 82.89 ($\pm$0.84) & 4.18 ($\pm$0.84) & 4.18 ($\pm$0.84) & 4.18 ($\pm$0.84) & 95.82 ($\pm$0.84) & 4.59 ($\pm$1.01) & 4.59 ($\pm$1.01) & 4.59 ($\pm$1.01) & 95.41 ($\pm$1.01)\\
			SODA \cite{soda2024hudson} & CVPR'24 & 11.56 ($\pm$1.85) & 11.56 ($\pm$1.85) & 11.56 ($\pm$1.85) & 88.44 ($\pm$1.85) & 13.01 ($\pm$2.16) & 13.02 ($\pm$2.16) & 13.01 ($\pm$2.16) & 86.98 ($\pm$2.16) & 2.82 ($\pm$0.54) & 2.82 ($\pm$0.55) & 2.83 ($\pm$0.55) & 97.18 ($\pm$0.55) & 3.99 ($\pm$1.13) & 3.98 ($\pm$1.13) & 3.99 ($\pm$1.13) & 96.02 ($\pm$1.13)\\
			SODA* \cite{soda2024hudson} & CVPR'24 & 10.12 ($\pm$1.41) & 10.12 ($\pm$1.41) & 10.12 ($\pm$1.41) & 89.88 ($\pm$1.41) & 17.47 ($\pm$1.09) & 17.47 ($\pm$1.09) & 17.47 ($\pm$1.08) & 82.53 ($\pm$1.09) & 3.34 ($\pm$0.20) & 3.34 ($\pm$0.20) & 3.34 ($\pm$0.20) & 96.66 ($\pm$0.20) & 6.47 ($\pm$1.54) & 6.47 ($\pm$1.54) & 6.47 ($\pm$1.53) & 93.53 ($\pm$1.54)\\
			\emph{l}-DAE \cite{ldae2025chen} & ICLR'25 & 25.62 ($\pm$1.55) & 25.62 ($\pm$1.56) & 25.62 ($\pm$1.54) & 74.38 ($\pm$1.55) & 40.00 ($\pm$3.95) & 40.00 ($\pm$3.96) & 40.01 ($\pm$3.94) & 60.00 ($\pm$3.96) & 11.27 ($\pm$1.05) & 11.27 ($\pm$1.06) & 11.27 ($\pm$1.04) & 88.73 ($\pm$1.06) & 38.24 ($\pm$12.62) & 38.23 ($\pm$12.60) & 38.26 ($\pm$12.64) & 61.76 ($\pm$12.61)\\
			DBAE \cite{dbae2025kim} & ICLR'25 & 18.32 ($\pm$0.92) & 18.33 ($\pm$0.92) & 18.32 ($\pm$0.92) & 81.67 ($\pm$0.92) & 20.48 ($\pm$1.01) & 20.48 ($\pm$1.01) & 20.49 ($\pm$1.01) & 79.52 ($\pm$1.01) & 6.07 ($\pm$0.77) & 6.07 ($\pm$0.77) & 6.07 ($\pm$0.76) & 93.93 ($\pm$0.77) & 6.28 ($\pm$1.41) & 6.28 ($\pm$1.41) & 6.28 ($\pm$1.41) & 93.72 ($\pm$1.41)\\
			CAMD \cite{camd2026} & TCSVT'26 & 34.01 ($\pm$1.17) & 34.03 ($\pm$1.17) & 33.98 ($\pm$1.18) & 65.98 ($\pm$1.17) & 33.88 ($\pm$1.24) & 33.88 ($\pm$1.26) & 33.88 ($\pm$1.22) & 66.12 ($\pm$1.25) & 16.76 ($\pm$2.25) & 16.76 ($\pm$2.25) & 16.76 ($\pm$2.25) & 83.24 ($\pm$2.25) & 19.72 ($\pm$1.28) & 19.72 ($\pm$1.28) & 19.72 ($\pm$1.29) & 80.28 ($\pm$1.28)\\
			DIS \cite{residual2026zhang} & CVPR'26 & 23.68 ($\pm$1.86) & 23.68 ($\pm$1.86) & 23.68 ($\pm$1.85) & 76.32 ($\pm$1.86) & 29.61 ($\pm$0.88) & 29.60 ($\pm$0.88) & 29.62 ($\pm$0.88) & 70.39 ($\pm$0.88) & 12.38 ($\pm$1.77) & 12.38 ($\pm$1.77) & 12.38 ($\pm$1.77) & 87.62 ($\pm$1.77) & 16.41 ($\pm$1.38) & 16.41 ($\pm$1.39) & 16.41 ($\pm$1.38) & 83.59 ($\pm$1.38)\\
			%			\hline
			Baseline (\textbf{Ours}) & This work & 19.92 ($\pm$1.47) & 19.91 ($\pm$1.47) & 19.92 ($\pm$1.47) & 80.09 ($\pm$1.47) & 30.02 ($\pm$0.38) & 30.02 ($\pm$0.38) & 30.02 ($\pm$0.38) & 69.98 ($\pm$0.38) & 6.23 ($\pm$0.45) & 6.23 ($\pm$0.45) & 6.23 ($\pm$0.45) & 93.77 ($\pm$0.45) & 6.88 ($\pm$1.31) & 6.89 ($\pm$1.31) & 6.88 ($\pm$1.31) & 93.11 ($\pm$1.31)\\
			\rowcolor{lightgray}
			SigDiffuser (\textbf{Ours}) & This work & \textbf{9.56} ($\pm$1.13) & \textbf{9.56} ($\pm$1.13) & \textbf{9.56} ($\pm$1.13) & \textbf{90.44} ($\pm$1.13) & \textbf{13.02} ($\pm$0.61) & \textbf{13.02} ($\pm$0.61) & \textbf{13.01} ($\pm$0.61) & \textbf{86.98} ($\pm$0.61) & \textbf{1.85} ($\pm$0.46) & \textbf{1.85} ($\pm$0.46) & \textbf{1.84} ($\pm$0.46) & \textbf{98.15} ($\pm$0.46) & \textbf{3.13} ($\pm$0.33) & \textbf{3.13} ($\pm$0.33) & \textbf{3.13} ($\pm$0.33) & \textbf{96.87} ($\pm$0.33)\\
			\whline{1.05pt}
	\end{tabular}}
\end{table*}

\subsubsection{Evaluation Metric}
\label{sec::metric}
We employ four metrics: Equal Error Rate (EER), False Acceptance Rate (FAR), False Rejection Rate (FRR), and Accuracy (Acc). For fairness, we follow the most stringent evaluation setting from \cite{avn2022li,transosv2024li,msdn2021liu}, implementing an open-set verifier based on three realistic design choices. (1) Since HSV compares a query against one or more references, we adopt the more difficult yet realistic single-reference (1-vs-1) protocol, despite multiple references typically yielding higher accuracy. (2) We use a global threshold across all writers rather than a user-specific one for decision-making, enabling more reliable and objective assessment. (3) We evaluate under the writer-independent scenario, verifying signatures without user-specific training during enrollment. This better reflects real-world deployment than writer-dependent methods that train personalized classifiers per writer. Implementation details are presented below.

Specifically, given a query signature $x^q$ claiming to be writer $w$ and a reference signature $x^w$ of writer $w$, we extract their 512-dimensional feature vectors $f_q = \phi(x_q)$ and $f_w = \phi(x_w)$ using the encoder and pooling layer $\phi$ of our downstream model, and measure their similarity by:
\begin{equation}
	s^w(x^q) = ||f_q - f_w||_2 / \sqrt{\overline{d_w}},
\end{equation}
where $||\cdot||_2$ refers to the Euclidean distance, $\overline{d_w}$ denotes the mean reference-to-reference distance for writer $w$, following \cite{synsig2vec2022lai}; in our 1-vs-1 setting with a single reference, $\overline{d_w} = 1$. Given a threshold $c$, the decision rule is defined as:
\begin{equation}
	\label{eq::decision}
	\delta_c(x^q, x^w) = \mathbbm{1}\!\left[\, s^w(x^q) < c \,\right],
\end{equation}
where $\mathbbm{1}[\cdot]$ is the indicator function; $\delta_c = 1$ accepts $x^q$ as genuine, and $\delta_c = 0$ rejects it as a forgery.

Consider a dataset comprising $W$ writers, each possessing $N$ genuine signatures and $M$ forged signatures. To achieve 1-vs-1 evaluation, we designate each genuine sample in turn as the reference while treating the remaining $N-1$ genuine signatures and $M$ forged signatures as queries. Since the comparison score is symmetric, each genuine-genuine pair is counted once, yielding $N \times (N-1) / 2$ genuine-genuine scores and $N \times M$ genuine-forged scores per writer. We then pool all the comparison scores from all writers into a single global evaluation set, resulting in $W \times (N \times (N-1) / 2 +N \times M)$ scores. Given a global threshold $c$, each score is converted into a binary decision according to Eq.~\ref{eq::decision}. We compare these decisions with the corresponding pair labels and count the numbers of True Positive (TP), False Positive (FP), True Negative (TN), and False Negative (FN). The FAR, FRR, and Acc at threshold $c$ are computed as:
\begin{equation}
	\begin{gathered}
		FAR(c) = \frac{FP(c)}{TN(c) + FP(c)},\\
		FRR(c) = \frac{FN(c)}{TP(c) + FN(c)},\\
		Acc(c) = \frac{TP(c) + TN(c)}{TP(c) + TN(c) + FP(c) + FN(c)}.
	\end{gathered}
\end{equation}

EER is the error rate at the threshold where FAR equals FRR. Let $\mathcal{C}$ denote the candidate threshold set consisting of all distinct comparison scores. Given the discrete and finite nature of $\mathcal{C}$, an exact threshold satisfying $FAR(c) = FRR(c)$ generally does not exist within $\mathcal{C}$. We therefore first search the candidate set $\mathcal{C}$ for a discrete threshold $c^*$ that minimizes the gap between FAR and FRR:
\begin{equation}
	c^* = \arg\min\limits_{c \in \mathcal{C}}|FAR(c) - FRR(c)|.
\end{equation}
Since the computation of FAR, FRR, and Acc requires a valid confusion matrix at an actual discrete threshold, we report them at this determined operating point:
\begin{equation}
	\text{FAR} = FAR(c^*),\text{FRR} = FRR(c^*),\text{Acc} = Acc(c^*).
\end{equation}

Subsequently, we compute EER at an exact threshold obtained through linear interpolation. While the discrete approximation $\tfrac{1}{2}(FAR(c^*) + FRR(c^*))$ can provide a direct estimate, it may deviate from the EER at the true FAR-FRR intersection. To obtain a precise EER, we linearly interpolate between $c^*$ and its adjacent threshold $c^\dagger \in \mathcal{C}$ lying on the opposite side of the FAR–FRR crossing to locate the crossing point, \emph{i.e.}, the exact threshold. Let $D(c) = FAR(c) - FRR(c)$, this threshold $\tilde{c}$ is given by:
\begin{equation}
	\tilde{c} = c^* + \frac{D(c^*)}{D(c^*) - D(c^\dagger)} \, (c^\dagger - c^*).
\end{equation}
At this threshold, the FAR and FRR equalize, allowing us to compute the EER as:
\begin{equation}
	\text{EER} = FAR(\tilde{c}) = FRR(\tilde{c}).
\end{equation}
Note that when $c^* = \tilde{c}$, $\text{EER}=\text{FAR}=\text{FRR}=1-\text{Acc}$.

Importantly, $c^*$ (and consequently $\tilde{c}$) is derived from the score distribution pooled across all writers, representing a global threshold rather than a writer-specific one. This implementation also accommodates an unlimited number of test writers, ensuring writer independence and generalization to unseen writers. All metrics are reported as percentages.

\subsection{Implementation Detail}
\label{sec::impl}
The maximum time step $T$ for denoiser sampling is set to 1000. $\alpha$, $\sigma$, and $K$ in RFR (Eq.~\ref{eq::alpha} and~\ref{eq::theta}) are set to $18,072$, $2.048 \times 10^6$, $91$, respectively. During pretraining, the side length $l$ for patches $I_{org}$, $I_{cor}$, and $I_{sty}$ is 96. During fine-tuning, all datasets use $96\times336$ (height $\times$ width) image inputs. More data processing details are included in Appendix A.

For pretraining, SigDiffuser is trained from scratch for 800 epochs (20 warmup epochs) using AdamW ($\beta_1=0.9$, $\beta_2=0.999$, and weight decay 0.05), a batch size of 32, and a learning rate of 1e-4. The additive noise term $\boldsymbol{\epsilon}$ in the denoiser follows a cosine schedule. For downstream training, we train the model for 80 epochs with a batch size of 24, using AdamW with a weight decay of 1e-5 and a cosine learning rate decay from 1e-4 to 1e-7. The baseline is a randomly initialized downstream model fine-tuned per dataset. Data augmentation is limited to random translation ($\le$5 pixels, padded with white). Training time scales with the dataset size. For example, on the largest HanSig dataset, pretraining consumes around four days and fine-tuning takes around two hours, respectively, on a single NVIDIA 3090 GPU. For standard data splits, we evaluate the final epoch's checkpoint. For splits with validation, we apply early stopping with a patience of 20 epochs and evaluate the best checkpoint.

\begin{table*}[t]
	\renewcommand{\arraystretch}{1.0}
	\centering
	\setlength{\tabcolsep}{4pt}
	\begin{minipage}{0.48\linewidth}
		\caption{Performance comparison on BHSig260-B \cite{bhsig2016pal}.}
		\label{Table::comp_sota_bengali}
		\centering
		\resizebox{\linewidth}{!}{
			\begin{tabular}{l c c c c c}
				\whline{1.05pt}
				\noalign{\vspace{2pt}}
				Method & Venue & EER $\downarrow$ & FAR $\downarrow$ & FRR $\downarrow$ & Acc $\uparrow$\\
				\hline
				SigNet \cite{signet2017dey} & arXiv'17 & 13.89 & 13.89 & 13.89 & 86.11\\
				%			\hline
				SigNet-F \cite{signetf2017hafemen} & PR'17 & 20.44 & 20.30 & 20.87 & 79.45\\
				%			\hline
				FMPNet \cite{pdsn2018lai} & ICFHR'18 & 13.36 & 13.33 & 13.37 & 86.65\\
				DeepHSV \cite{deephsv2019li} & ICDAR'19 & 11.92 & - & - & 88.08\\
				%			\hline
				IDN \cite{idn2019wei} & CVPR'19 & - & 4.12 & 5.24 & {95.32}\\
				%			\hline
				SDINet \cite{sdinet2021li} & AAAI'21 & - & 3.30 & 7.86 & 94.42\\
	%			MSN \cite{msn2021icdar} & ICDAR'21 & 8.43 & 10.42 & 6.44 & 91.56\\
				CaP \cite{cap2021lu} & ICPR'21 & {3.96} & {3.96} & {3.96} & -\\
				%			\hline
				AVN \cite{avn2022li} & TMM'22 & 6.14 & - & - & 93.80\\
				%			\hline
				2C2S \cite{2c2s2023ren} & EAAI'23 & 6.75 & 5.37 & 8.11 & 93.25\\
				MGRNet \cite{hansig2023huang} & arXiv'23 & 6.12 & 5.93 & 6.20 & -\\
				%			\hline
				LwR \cite{lwr2023ji} & PR'23 & 6.10 & - & - & 94.50\\
				TransOSV \cite{transosv2024li} & PR'24 & 9.90 & 9.90 & 9.90 & -\\
				SPD Manifold \cite{spd2024zois} & TIFS'24 & 11.10 & - & - & -\\
				%			\hline
				CWA \cite{cwa2024ling} & ICDAR'24 & \underline{2.53} & \underline{2.95} & \underline{2.14} & \underline{97.27}\\
				DetailSemNet \cite{detailsemnet2024eccv} & ECCV'24 & 10.19 & 5.96 & 18.24 & 90.64\\
				HTCSigNet \cite{htcsignet2025zheng} & PR'25 & 8.52 & 9.20 & 7.83 & 91.17\\ 
				\hline
				\noalign{\vspace{1pt}}
				Baseline (\textbf{Ours}) & This work & 6.23 ($\pm$0.45) & 6.23 ($\pm$0.45) & 6.23 ($\pm$0.45) & 93.77 ($\pm$0.45)\\
				%			\hline
				SigDiffuser$^\text{val}$ (\textbf{Ours}) & This work & 2.23 ($\pm$0.53) & 2.23 ($\pm$0.53) & 2.23 ($\pm$0.53) & 97.77 ($\pm$0.53)\\
				\rowcolor{lightgray}
				SigDiffuser (\textbf{Ours}) & This work & \textbf{1.85} ($\pm$0.46) & \textbf{1.85} ($\pm$0.46) & \textbf{1.84} ($\pm$0.46) & \textbf{98.15} ($\pm$0.46)\\
				\whline{1.05pt}
		\end{tabular}}
	\end{minipage}
	\hfill
	\hspace{-0.6em}
	\begin{minipage}{0.48\linewidth}
		\caption{Performance comparison on BHSig260-H \cite{bhsig2016pal}.}
		\label{Table::comp_sota_hindi}
		\centering
		\resizebox{\linewidth}{!}{
			\begin{tabular}{l c c c c c}
				\whline{1.05pt}
				\noalign{\vspace{2pt}}
				Method & Venue & EER $\downarrow$ & FAR $\downarrow$ & FRR $\downarrow$ & Acc $\uparrow$\\
				\hline
				SigNet \cite{signet2017dey} & arXiv'17 & 15.36 & 15.36 & 15.36 & 84.64\\
				%			\hline
				SigNet-F \cite{signetf2017hafemen} & PR'17 & 18.30 & 18.22 & 18.37 & 81.71\\
				%			\hline
				FMPNet \cite{pdsn2018lai} & ICFHR'18 & 14.28 & 14.17 & 14.48 & 85.70\\
				DeepHSV \cite{deephsv2019li} & ICDAR'19 & 13.34 & - & - & 86.66\\
				%			\hline
				IDN \cite{idn2019wei} & CVPR'19 & - & 8.99 & 4.93 & 93.04\\
				SDINet \cite{sdinet2021li} & AAAI'21 & - & 6.24 & 3.77 & \underline{95.00}\\
				%			\hline
	%			MSN \cite{msn2021icdar} & ICDAR'21 & 11.31 & 17.06 & 5.16 & 88.88\\
				CaP \cite{cap2021lu} & ICPR'21 & 5.97 & 5.97 & 5.97 & -\\
				%			\hline
				AVN \cite{avn2022li} & TMM'22 & 5.65 & - & - & {94.32}\\
				%			\hline
				2C2S \cite{2c2s2023ren} & EAAI'23 & 9.32 & 8.66 & 9.98 & 90.68\\
				MGRNet \cite{hansig2023huang} & arXiv'23 & 6.68 & 6.76 & 6.56 & -\\
				%			\hline
				LwR \cite{lwr2023ji} & PR'23 & 8.80 & - & - & 92.60\\
				TransOSV \cite{transosv2024li} & PR'24 & \underline{3.24} & \underline{3.24} & \underline{3.24} & -\\
				SPD Manifold \cite{spd2024zois} & TIFS'24 & 15.60 & - & - & -\\
				%			\hline
				CWA \cite{cwa2024ling} & ICDAR'24 & 9.55 & 7.11 & 12.14 & 91.42\\
				DetailSemNet \cite{detailsemnet2024eccv} & ECCV'24 & 13.31 & 7.89 & 22.32 & 88.11\\
				HTCSigNet \cite{htcsignet2025zheng} & PR'25 & 4.63 & 4.61 & 4.65 & 95.26\\
				\hline
				\noalign{\vspace{1pt}}
				Baseline (\textbf{Ours}) & This work & 6.88 ($\pm$1.31) & 6.89 ($\pm$1.31) & 6.88 ($\pm$1.31) & 93.11 ($\pm$1.31)\\
				%			\hline
				SigDiffuser$^\text{val}$ (\textbf{Ours}) & This work & 3.65 ($\pm$0.44) & 3.65 ($\pm$0.44) & 3.65 ($\pm$0.44) & 96.35 ($\pm$0.44)\\
				\rowcolor{lightgray}
				SigDiffuser (\textbf{Ours}) & This work & \textbf{3.13} ($\pm$0.33) & \textbf{3.13} ($\pm$0.33) & \textbf{3.13} ($\pm$0.33) & \textbf{96.87} ($\pm$0.33)\\
				\whline{1.05pt}
		\end{tabular}}
	\end{minipage}
\end{table*}

\begin{table*}[t]
	\begin{minipage}[t]{0.48\linewidth}
		\caption{Performance comparison on HanSig \cite{hansig2023huang}.}
		\label{Table::comp_sota_hansig}
		\centering
		\setlength{\tabcolsep}{4pt}
		\renewcommand{\arraystretch}{1.1}
		\resizebox{\linewidth}{!}{
			\begin{tabular}{l c c c c c}
				\whline{1.05pt}
				\noalign{\vspace{2pt}}
				Method & Venue & EER $\downarrow$ & FAR $\downarrow$ & FRR $\downarrow$ & Acc $\uparrow$\\
				\hline
				SigNet \cite{signet2017dey} & arXiv'17 & 33.07 & 33.10 & 33.05 & 66.92\\
				%			\hline
				SigNet-F \cite{signetf2017hafemen} & PR'17 & 41.71 & 41.80 & 41.58 & 58.31\\
				%			\hline
				FMPNet \cite{pdsn2018lai} & ICFHR'18 & 13.13 & 13.10 & 13.16 & 86.87\\
				%			\hline
				MGRNet \cite{hansig2023huang} & arXiv'23 & {9.93} & 11.85 & \underline{7.69} & -\\
				%			\hline
				CWA \cite{cwa2024ling} & ICDAR'24 & \underline{9.47} & \textbf{6.61} & 16.23 & 90.26\\
				DetailSemNet \cite{detailsemnet2024eccv} & ECCV'24 & 16.00 & 10.19 & 24.62 & 85.17\\
				HTCSigNet \cite{htcsignet2025zheng} & PR'25 & 22.18 & 22.18 & 22.18 & 77.82\\
				\hline
				\noalign{\vspace{1pt}}
				Baseline (\textbf{Ours}) & This work & 19.22 ($\pm$8.99) & 19.22 ($\pm$8.98) & 19.23 ($\pm$8.99) & 80.78 ($\pm$8.99)\\
				%			\hline
				SigDiffuser$^\text{val}$ (\textbf{Ours}) & This work & 7.69 ($\pm$2.23) & 7.69 ($\pm$2.23) & 7.69 ($\pm$2.24) & 92.31 ($\pm$2.23)\\
				\rowcolor{lightgray}
				SigDiffuser (\textbf{Ours}) & This work & \textbf{7.39} ($\pm$1.64) & \underline{7.39} ($\pm$1.63) & \textbf{7.39} ($\pm$1.64) & \textbf{92.61} ($\pm$1.64)\\
				\whline{1.05pt}
		\end{tabular}}
	\end{minipage}
	\hfill
	\hspace{-0.6em}
	\begin{minipage}[t]{0.48\linewidth}
		\caption{Performance comparison on UTSig \cite{utsig2017soleimani}.}
		\label{Table::comp_sota_utsig}
		\centering
		\setlength{\tabcolsep}{2.7pt}
		\renewcommand{\arraystretch}{1.1}
		\resizebox{\linewidth}{!}{
			\begin{tabular}{l c c c c c}
				\whline{1.05pt}
				\noalign{\vspace{2pt}}
				Method & Venue & EER $\downarrow$ & FAR $\downarrow$ & FRR $\downarrow$ & Acc $\uparrow$\\
				\hline
				\noalign{\vspace{1pt}}
				UTSig-S1 \cite{utsig2017soleimani} & IET Biom'16 & 29.71 & \underline{21.29} & 39.27 & -\\
				%			\hline
				SigNet \cite{signet2017dey} & arXiv'17 & 47.08 & 46.71 & 47.38 & 53.05\\
				%			\hline
				SigNet-F \cite{signetf2017hafemen} & PR'17 & 43.20 & 42.93 & 43.64 & 56.81\\
				%			\hline
				FMPNet \cite{pdsn2018lai} & ICFHR'18 & {32.95} & 32.81 & 33.04 & {67.11}\\
				%			\hline
				LwR \cite{lwr2023ji} & PR'23 & {29.70} & - & - & \underline{72.70}\\
				DetailSemNet \cite{detailsemnet2024eccv} & ECCV'24 & 39.54 & \textbf{0.78} & 96.31 & 77.80\\
				HTCSigNet \cite{htcsignet2025zheng} & PR'25 & \underline{28.93} & 28.92 & \underline{28.96} & 71.08\\
				\hline
				\noalign{\vspace{1pt}}
				Baseline (\textbf{Ours}) & This work & 29.81 ($\pm$1.51) & 29.81 ($\pm$1.51) & 29.81 ($\pm$1.51) & 70.19 ($\pm$1.51)\\ 
				%			\hline
				SigDiffuser$^\text{val}$ (\textbf{Ours}) & This work & 27.39 ($\pm$1.62) & 27.38 ($\pm$1.61) & 27.40 ($\pm$1.62) & 72.61 ($\pm$1.62)\\
				\rowcolor{lightgray}
				SigDiffuser (\textbf{Ours}) & This work & \textbf{23.48} ($\pm$1.81) & {23.48} ($\pm$1.81) & \textbf{23.49} ($\pm$1.82) & \textbf{76.52} ($\pm$1.81)\\
				\whline{1.05pt}
		\end{tabular}}
	\end{minipage}
\end{table*}

\begin{table}[t]
	\renewcommand{\arraystretch}{1.0}
	\caption{Performance comparison on CEDAR \cite{cedar2004kalera}.}
	\label{Table::comp_sota_cedar}
	\centering
	\setlength{\tabcolsep}{3.5pt}
	\resizebox{0.98\linewidth}{!}{
		\begin{tabular}{l c c c c c}
			\whline{1.05pt}
			\noalign{\vspace{2pt}}
			Method & Venue & EER $\downarrow$ & FAR $\downarrow$ & FRR $\downarrow$ & Acc $\uparrow$\\
			\hline
			SigNet-F \cite{signetf2017hafemen} & PR'17 & 5.92 & - & - & -\\
			FMPNet \cite{pdsn2018lai} & ICFHR'18 & 12.17 & 12.08 & 12.17 & 87.87\\
			IDN \cite{idn2019wei} & CVPR'19 & 3.62 & 5.87 & \textbf{2.17} & -\\
			SDINet \cite{sdinet2021li} & AAAI'21 & \textbf{1.75} & \textbf{0.73} & {3.42} & -\\
			MSDN \cite{msdn2021liu} & PR'21 & 6.74 & - & - & -\\
			CaP \cite{cap2021lu} & ICPR'21 & 4.34 & 4.34 & 4.34 & -\\
			ILNNeck \cite{domainshift2022ijcnn} & IJCNN'22 & 9.21 & - & - & \underline{90.79}\\
			QSLS \cite{qsls2023ajij} & Neural Comput. Appl.'23 & \underline{1.93} & 3.35 & \underline{2.39} & -\\ 
			MGRNet \cite{hansig2023huang} & arXiv'23 & 3.51 & \underline{3.33} & 3.55 & -\\
			SPD Manifold \cite{spd2024zois} & TIFS'24 & 8.53 & - & - & -\\
			DetailSemNet \cite{detailsemnet2024eccv} & ECCV'24 & 24.61 & 10.97 & 40.29 & 79.53\\
			DCCCAE \cite{dcccae2025zhao} & PR'25 & 10.81 & - & - & -\\
			\hline
			\noalign{\vspace{1pt}}
			Baseline (\textbf{Ours}) & This work & 12.77 ($\pm$8.42) & 12.77 ($\pm$8.42) & 12.75 ($\pm$8.42) & 87.23 ($\pm$8.42)\\
			SigDiffuser$^\text{val}$ (\textbf{Ours}) & This work & 8.55 ($\pm$6.91) & 8.55 ($\pm$6.91) & 8.54 ($\pm$6.91) & 91.45 ($\pm$6.91)\\
			\rowcolor{lightgray}
			SigDiffuser (\textbf{Ours}) & This work & 6.92 ($\pm$5.88) & 6.92 ($\pm$5.87) & 6.92 ($\pm$5.88) & \textbf{93.08} ($\pm$5.88)\\
			\whline{1.05pt}
	\end{tabular}}
\end{table}

\begin{table*}[t]
	\renewcommand{\arraystretch}{0.95}
	\caption{Ablation study on BHSig260-B \cite{bhsig2016pal} and BHSig260-H \cite{bhsig2016pal}. SSRL denotes self-supervised representation learning.}
	\label{Table::ablation_arch}
	\centering
	\resizebox{0.95\textwidth}{!}{
		\begin{tabular}{c c c c c c c c c c}
			\toprule
			\multirow{2.4}{*}{\#Line} & \multirow{2.4}{*}{Case} & \multicolumn{4}{c}{BHSig260-B} & \multicolumn{4}{c}{BHSig260-H}\\
			\cmidrule(r){3-6}\cmidrule(r){7-10}
			~ & ~ & EER $\downarrow$ & FAR $\downarrow$ & FRR $\downarrow$ & Acc $\uparrow$ & EER $\downarrow$ & FAR $\downarrow$ & FRR $\downarrow$ & Acc $\uparrow$\\
			\hline
			\noalign{\vspace{1pt}}
			1 & Baseline & 5.77 & 5.78 & 5.76 & 94.23 & 5.95 & 5.96 & 5.95 & 94.04\\
			2 & Baseline w/ RFR-based SSRL; w/ ${I}_{cor}$ & 2.60 & 2.60 & 2.60 & 97.40 & 5.18 & 5.19 & 5.18 & 94.82\\
			\rowcolor{lightgray}
			3 & Baseline w/ RFR-based SSRL; w/ ${I}_{cor}$; w/ PFAM & \textbf{1.58} & \textbf{1.58} & \textbf{1.57} & \textbf{98.43} & \textbf{2.59} & \textbf{2.59} & \textbf{2.58} & \textbf{97.41}\\
			4 & Baseline w/ RFR-based SSRL; w/o ${I}_{cor}$; w/ PFAM & 4.34 & 4.34 & 4.33 & 95.66 & 3.31 & 3.31 & 3.31 & 96.69\\
			5 & Baseline w/ MIM-based SSRL; w/ $I_{cor}$; w/ PFAM & 3.39 & 3.39 & 3.39 & 96.61 & 4.13 & 4.13 & 4.13 & 95.87\\
			\noalign{\vspace{-2pt}}
			\bottomrule
	\end{tabular}}
\end{table*}

\subsection{Linear Probing and Downstream Fine-Tuning Evaluation}
\label{sec::probe_and_full}
We investigate SigDiffuser's representation quality through probe evaluation (freezing the encoder, training only NSPP and PDSN) and downstream fine-tuning (full-parameter training). Baselines include contrastive methods (\emph{e.g.}, SimCLR \cite{simclr2020icml}, SeqCLR \cite{seqclr2021aberdam}), MIM-based generative methods (\emph{e.g.}, MAE \cite{mae2022he}, SparK \cite{spark2023tian}, CAMD \cite{camd2026}), diffusion-based generative methods (\emph{e.g.}, DiffAE \cite{diffae2022preechakul}, SODA \cite{soda2024hudson}, \emph{l}-DAE \cite{ldae2025chen}). For fair comparison, most baselines are reimplemented using SigDiffuser's ResNet-18 encoder. Exceptions retaining their native architectures are MAE, CAMD, and DIS (ViT-Tiny \cite{vit2021dosovitskiy}), and \emph{l}-DAE (DiT-S \cite{dit2023peebles}). Baselines evaluated with their original encoders are marked with *, $^\dagger$, and $^\star$.

Results are summarized in Table~\ref{Table::finetune}. (1) SigDiffuser consistently achieves the best performance in both probe and fine-tuning evaluations. Compared to the randomly initialized baseline, the pretrained model achieves superior mean performance and generally lower across-split variance, demonstrating that our pretraining extracts high-quality representations, directly improving verification stability. (2) Most competing methods, even those with pretext tasks tailored for handwritten/scene text self-supervised representation learning (SSRL) \cite{seqclr2021aberdam,chaco2022zhang,cmtco2022zhang,siman2022luo}, perform significantly worse, exhibiting lower means and higher across-split variance. SigDiffuser also surpasses other diffusion-based schemes, such as DiffAE and SODA, by considerable margins. (3) Models using their original encoders, whether deeper (ResNet-50), modern-designed (ConvNeXt-Small), or lightweight (VGG-10), consistently underperform our default ResNet-18. Although native architectures align better with their vanilla designs, these results indicate that ResNet-18 provides a superior balance of capacity and architectural suitability for SSRL in HSV.

(4) Generative methods generally outperform contrastive methods, suggesting contrastive learning is less suited for HSV. Among generative methods, diffusion-based approaches mostly surpass MIM-based models, consistent with the blurred reconstruction results shown in Fig.~\ref{Fig::comp} (a). This is potentially due to MIM's difficulty in perceiving fine textures amidst white backgrounds when large portions of strokes are masked. (5) SigDiffuser's superiority mainly stems from the proposed RFR. Unlike pretext tasks with mild deformations that permit local pixel-copying shortcuts, RFR requires the model to restore a severely ruined signature region guided by a spatially distinct style patch. As discussed in Sec.~\ref{sec::pretext}, this design suppresses reconstruction shortcuts and forces reliance on intrinsic writing-style evidence encoded from $I_{sty}$. Hence, the encoder captures writing-style cues across spatial locations (\emph{e.g.}, stroke dynamics, spacing, and slant) rather than trivial pixel relations, which are essential for HSV. Further visual analyses are presented in Sec.~\ref{sec::repr_vis}.

\subsection{Comparison with Existing HSV Method}
We compare the full-parameter fine-tuned model with SOTA HSV methods in Tables~\ref{Table::comp_sota_bengali} to \ref{Table::comp_sota_utsig}. For fair comparison, we only include prior results obtained under identical training/testing splits and evaluation settings. For instance, we report single-reference rather than multi-reference results for HTCSigNet \cite{htcsignet2025zheng} and the 50/50 writer split for TransOSV \cite{transosv2024li} on BHSig260-B (which also reports 80/20). Here we report the multi-split results, with per-split details provided in Appendix B.

As confirmed by the results, even under the multi-split protocol, SigDiffuser outperforms preceding models in almost all cases. On BHSig260-B and BHSig260-H (Tables~\ref{Table::comp_sota_bengali} and~\ref{Table::comp_sota_hindi}), SigDiffuser sets new records by improving EERs from 2.53\% and 3.24\% to 1.85\% and 3.13\%, respectively. Although the second-best method, CWA, uses an 80/20 writer split (easier than our 50/50 split), SigDiffuser maintains its superiority. HanSig is a challenging Chinese HSV dataset characterized by significant variation in writing style. Despite the heightened difficulty, SigDiffuser outperforms all existing methods in Table~\ref{Table::comp_sota_hansig}. On UTSig (Table~\ref{Table::comp_sota_utsig}), SigDiffuser shows significant advantages over existing methods, even approaching the performance of 12-reference methods despite using only one reference. It is noteworthy that the signatures across different datasets diverge drastically in text structure and writing styles due to language and cultural discrepancies. Overall, the strong average results across multiple data splits demonstrate SigDiffuser's effectiveness in learning informative writing patterns that transcend linguistic and stylistic boundaries.

We further evaluate on the CEDAR dataset. Several prior works \cite{2c2s2023ren,deephsv2019li,signet2017dey,cwa2024ling} report seemingly perfect metrics (0\% EER and 100\% Acc) by relying on trivial background cues (genuine signatures are gray, forgeries are white) rather than actual strokes. To prevent this, we remove all backgrounds during data preprocessing. Table~\ref{Table::comp_sota_cedar} shows that RFR pretraining substantially improves SigDiffuser over the baseline. Notably, although the inherent difficulty of certain writers and samples in CEDAR causes high performance variance, our pretraining effectively reduces this instability ($\pm6.91$$\rightarrow$$\pm5.88$). While not achieving the lowest EER, SigDiffuser remains competitive.

Other than standard data splits, we also evaluate using splits with validation (Sec.~\ref{sec::dataset}), denoted as SigDiffuser$^\text{val}$. As observed, performance consistently drops across all datasets. We attribute this degradation to the reduced training data volume. Since offline HSV datasets are inherently small, mainstream methods typically use only training-testing splits to maximize training data; introducing a validation set inevitably hampers performance. Despite this decline, SigDiffuser$^\text{val}$ maintains optimal or near-optimal performance on most datasets.

\subsection{Ablation Study}
We analyze the effect of SigDiffuser's key components, including the RFR pretext task, the pyramidal feature aggregation module (PFAM), and the corrupted signature input $I_{cor}$, on a single data split, with results presented in Table~\ref{Table::ablation_arch}. Comparing Lines 1 and 2, self-supervised learning with RFR effectively improves HSV performance. In Line 3, adding PFAM to form the complete SigDiffuser further improves EER by 1.02\% and 2.59\%, confirming PFAM's efficacy.

Additionally, since using $I_{sty}$ and $I_{org}$ also enables representation learning, we investigate the efficacy and necessity of introducing the corrupted signature region $I_{cor}$. Conceptually, removing $I_{cor}$ degrades our RFR pretext task to novel view synthesis. As shown in Lines 3 and 4, removing $I_{cor}$ leads to significant performance declines, especially on BHSig260-B. This suggests that reconstructing from the corrupted signatures $I_{cor}$ is effective to boost model performance, underscoring its necessity as RFR's core component. Furthermore, substituting RFR with MIM drops overall performance (Lines 3 vs. 5). This indicates RFR's superiority over MIM in HSV modeling, aligning with our findings in Sec.~\ref{sec::probe_and_full}. Additional ablation studies regarding PFAM's implementation variants, the isolated effect of diffusion modeling in SigDiffuser, the number of pretraining epochs, and patch side length $l$ used in RFR are detailed in Appendices C, D, E, and F, respectively.

\begin{table*}[t]
	\renewcommand{\arraystretch}{0.9}
	\centering
	\caption{Cross-dataset validation. Models are trained on the source dataset and tested on the target dataset. $\Delta_{\text{Acc}}$ denotes the accuracy drop from the best in-domain accuracy, where smaller values indicate better cross-dataset generalization.}
	\label{Table::cross}
	\newcommand{\dg}[1]{\cellcolor{gray!8}#1}
	\newcommand{\bestdg}[1]{\cellcolor{gray!8}\textbf{#1}}
	\resizebox{\linewidth}{!}{
		\begin{tabular}{ll c c c c c c c c c c  c c c}
			\toprule
			\noalign{\vspace{-1.5pt}}
			\multirow{3.6}{*}{Source} & \multirow{3.6}{*}{Model} & \multirow{3.6}{*}{Venue} & \multicolumn{12}{c}{Target}\\
			\noalign{\vspace{-1pt}}
			\cmidrule(r){4-6}\cmidrule(r){7-9}\cmidrule(r){10-12}\cmidrule(r){13-15}
			\noalign{\vspace{-1pt}}
			~ & ~ & ~ & \multicolumn{3}{c}{BHSig260-B} & \multicolumn{3}{c}{BHSig260-H} & \multicolumn{3}{c}{CEDAR} & \multicolumn{3}{c}{HanSig}\\
			\noalign{\vspace{-1pt}}
			\cmidrule(r){4-6}\cmidrule(r){7-9}\cmidrule(r){10-12}\cmidrule(r){13-15}
			\noalign{\vspace{-1pt}}
			~ & ~ & ~ & EER $\downarrow$ & Acc $\uparrow$ & {$\Delta_\text{Acc} \downarrow$} & EER $\downarrow$ & Acc $\uparrow$ & {$\Delta_\text{Acc} \downarrow$} & EER $\downarrow$ & Acc $\uparrow$ & {$\Delta_\text{Acc} \downarrow$} & EER $\downarrow$ & Acc $\uparrow$ & {$\Delta_\text{Acc} \downarrow$}\\
			\hline
			\noalign{\vspace{2pt}}
			\multirow{7}{*}{BHSig260-B} & SigNet \cite{signet2017dey} & arXiv'17 & 33.89 & 86.11 & \dg{0.00} & - & 52.78 & \dg{33.33} & - & 50.00 & \dg{36.11} & - & - & \dg{-}\\
			~ & IDN \cite{idn2019wei} & CVPR'19 & - & 95.32 & \dg{0.00} & - & 74.30 & \dg{21.02} & - & 50.00 & \dg{45.32} & - & - & \dg{-}\\
			~ & CaP \cite{cap2021lu} & ICDAR'21 & 3.96 & 96.04 & \dg{0.00} & - & 69.59 & \dg{26.45} & - & 66.29 & \dg{29.75} & - & - & \dg{-}\\
			~ & ILNNeck \cite{domainshift2022ijcnn} & IJCNN'22 & 5.72 & 94.28 & \dg{0.00} & - & 79.21 & \dg{15.07} & - & 80.34 & \bestdg{13.94} & - & - & \dg{-}\\
			~ & DetailSemNet \cite{detailsemnet2024eccv} & ECCV'24 & 10.19 & 90.64 & \dg{0.00} & 26.00 & 78.69 & \dg{11.95} & 33.40 & 72.35 & \dg{18.29} & 36.01 & 70.41 & \dg{20.23}\\
			~ & Baseline (\textbf{Ours}) & This work & 5.77 & 94.23 & \dg{0.00} & 22.69 & 77.32 & \dg{16.91} & 28.85 & 71.15 & \dg{23.08} & 24.70 & 75.31 & \dg{18.92}\\
			~ & SigDiffuser (\textbf{Ours}) & This work & {1.58} & {98.43} & \dg{0.00} & {12.44} & {87.56} & \bestdg{10.87} & 16.95 & 83.05 & \dg{15.38} & 19.06 & 80.94 & \bestdg{17.49}\\
			\midrule
			\multirow{7}{*}{BHSig260-H} & SigNet & arXiv'17 & - & 60.65 & \dg{23.99} & 15.36 & 84.64 & \dg{0.00} & - & 59.57 & \dg{25.07} & - & - & \dg{-}\\
			~ & IDN & CVPR'19 & - & 74.12 & \dg{18.92} & - & 93.04 & \dg{0.00} & - & 50.00 & \dg{43.04} & - & - & \dg{-}\\
			~ & CaP & ICDAR'21 & - & 85.34 & \dg{8.69} & 5.97 & 94.03 & \dg{0.00} & - & 70.51 & \dg{23.52} & - & - & \dg{-}\\
			~ & ILNNeck & IJCNN'22 & - & 87.57 & \dg{8.22} & 4.21 & 95.79 & \dg{0.00} & - & 79.76 & \dg{16.03} & - & - & \dg{-}\\
			~ & DetailSemNet & ECCV'24 & 22.07 & 80.52 & \dg{7.59} & 13.31 & 88.11 & \dg{0.00} & 35.94 & 71.13 & \dg{16.98} & 36.46 & 69.72 & \dg{18.39}\\
			~ & Baseline (\textbf{Ours}) & This work & 11.27 & 88.73 & \dg{5.31} & 5.95 & 94.04 & \dg{0.00} & 21.53 & 78.47 & \dg{15.57} & 24.93 & 75.07 & \dg{18.97}\\
			~ & SigDiffuser (\textbf{Ours}) & This work & 6.45 & 93.55 & \bestdg{3.86} & 2.59 & {97.41} & \dg{0.00} & 15.68 & 84.32 & \bestdg{13.09} & 20.37 & 79.63 & \bestdg{17.78}\\
			\midrule
			\multirow{6}{*}{CEDAR} & IDN & CVPR'19 & - & 50.01 & \dg{46.37} & - & 50.36 & \dg{46.02} & 3.62 & 96.38 & \dg{0.00} & - & - & \dg{-}\\
			~ & CaP & ICDAR'21 & - & 61.93 & \dg{33.63} & - & 60.92 & \dg{35.46} & 4.34 & 95.56 & \dg{0.00} & - & - & \dg{-}\\
			~ & ILNNeck & IJCNN'22 & - & 84.93 & \dg{5.86} & - & 80.83 & \dg{9.96} & 9.21 & 90.79 & \dg{0.00} & - & - & \dg{-}\\
			~ & DetailSemNet & ECCV'24 & 32.00 & 73.75 & \bestdg{5.78} & 33.67 & 73.40 & \bestdg{6.13} & 24.61 & 79.53 & \dg{0.00} & 39.71 & 68.19 & \bestdg{10.34}\\
			~ & Baseline (\textbf{Ours}) & This work & 32.54 & 67.46 & \dg{27.42} & 32.25 & 67.70 & \dg{27.18} & 5.11 & 94.88 & \dg{0.00} & 27.70 & 72.33 & \dg{22.55}\\
			~ & SigDiffuser (\textbf{Ours}) & This work & 21.08 & 78.92 & \dg{18.62} & 29.04 & 70.97 & \dg{26.57} & 2.46 & 97.54 & \dg{0.00} & 23.90 & 76.10 & \dg{21.44}\\
			\midrule
			\multirow{4}{*}{HanSig} & FMPNet \cite{pdsn2018lai} & ICFHR'18 & 23.16 & 76.83 & \dg{9.53} & 26.52 & 73.47 & \dg{12.37} & 22.25 & 77.75 & \dg{8.10} & 14.15 & 85.85 & \dg{0.00}\\
			~ & DetailSemNet & ECCV'24 & 20.85 & 80.72 & \dg{4.45} & 21.92 & 80.46 & \bestdg{4.71} & 22.54 & 81.01 & \dg{4.16} & 16.00 & 85.17 & \dg{0.00}\\
			~ & Baseline (\textbf{Ours}) & This work & 14.61 & 85.39 & \dg{4.53} & 30.18 & 69.82 & \dg{20.10} & 16.59 & 83.40 & \dg{6.52} & 10.07 & 89.92 & \dg{0.00}\\
			~ & SigDiffuser (\textbf{Ours}) & This work & 10.66 & 89.34 & \bestdg{3.30} & 25.99 & 74.02 & \dg{18.62} & 11.32 & 88.69 & \bestdg{3.95} & 7.36 & 92.64 & \dg{0.00}\\
			\noalign{\vspace{-2pt}}
			\bottomrule
	\end{tabular}}
\end{table*}

\begin{table*}[t]
	\centering
	\caption{Comparison of corruption degrees on BHSig260-B \cite{bhsig2016pal} and BHSig260-H \cite{bhsig2016pal}. (a) Setting 1. We use $\alpha$ to control the crumpling severity, ranging from 0 (no crumpling) to 27,108 (highly intensified). (b) Setting 2. We use existing data-augmentation techniques to control the crumpling degree: \emph{None} (SimAN \cite{siman2022luo}), \emph{Mild} (CMT-Co \cite{cmtco2022zhang}), and \emph{Full} (our RFR).}
	\label{Table::comp_corrupt}
	\begin{minipage}{0.48\textwidth}
		\centering
		\renewcommand{\arraystretch}{0.9}
		\label{Table::comp_corrupta}
		\subcaption{Setting 1: Controlled by $\alpha$ as described in Sec.~\ref{sec::pretext}.}
		\resizebox{\linewidth}{!}{
			\begin{tabular}{c c c c c | c c c c}
				\toprule
				\multirow{2.4}{*}{$\alpha$} & \multicolumn{4}{c}{BHSig260-B} & \multicolumn{4}{c}{BHSig260-H}\\
				\cmidrule(r){2-5}\cmidrule(r){6-9}
				~ & EER $\downarrow$ & FAR $\downarrow$ & FRR $\downarrow$ & Acc $\uparrow$ & EER $\downarrow$ & FAR $\downarrow$ & FRR $\downarrow$ & Acc $\uparrow$\\
				\hline
				\noalign{\vspace{2pt}}
				0 & 6.39 & 6.39 & 6.39 & 93.61 & 5.42 & 5.41 & 5.42 & 94.59\\
				9,036 & 1.88 & 1.88 & 1.88 & 98.12 & 3.21 & 3.21 & 3.21 & 96.79\\
				\rowcolor{lightgray}
				18,072 & \textbf{1.58} & \textbf{1.58} & \textbf{1.57} & \textbf{98.43} & \textbf{2.59} & \textbf{2.59} & \textbf{2.59} & \textbf{97.41}\\
				27,108 & 2.19 & 2.19 & 2.19 & 97.81 & 3.05 & 3.05 & 3.05 & 96.95\\
				\noalign{\vspace{-2pt}}
				\bottomrule
		\end{tabular}}		
	\end{minipage}
	\hfill
	\hspace{-2em}
	\begin{minipage}{0.48\textwidth}
		\centering
		\renewcommand{\arraystretch}{1.12}
		\label{Table::comp_corruptb}
		\subcaption{Setting 2: Controlled by existing implementations.}
		\resizebox{\linewidth}{!}{
			\begin{tabular}{c c c c c | c c c c}
				\toprule
				\multirow{2.4}{*}{Degree} & \multicolumn{4}{c}{BHSig260-B} & \multicolumn{4}{c}{BHSig260-H}\\
				\cmidrule(r){2-5}\cmidrule(r){6-9}
				~ & EER $\downarrow$ & FAR $\downarrow$ & FRR $\downarrow$ & Acc $\uparrow$ & EER $\downarrow$ & FAR $\downarrow$ & FRR $\downarrow$ & Acc $\uparrow$\\
				\hline
				\emph{None} & 4.59 & 4.59 & 4.59 & 95.41 & 5.73 & 5.73 & 5.72 & 94.27\\
%				\hline
				\emph{Mild} & 2.00 & 2.00 & 2.01 & 98.00 & 3.68 & 3.68 & 3.68 & 96.32\\
%				\hline
				\rowcolor{lightgray}
				\emph{Full} & \textbf{1.58} & \textbf{1.58} & \textbf{1.57} & \textbf{98.43} & \textbf{2.59} & \textbf{2.59} & \textbf{2.59} & \textbf{97.41}\\
				\noalign{\vspace{-2pt}}
				\bottomrule
		\end{tabular}}
	\end{minipage}
\end{table*}

\begin{figure}[t]
	\centering
	\includegraphics[width=\linewidth]{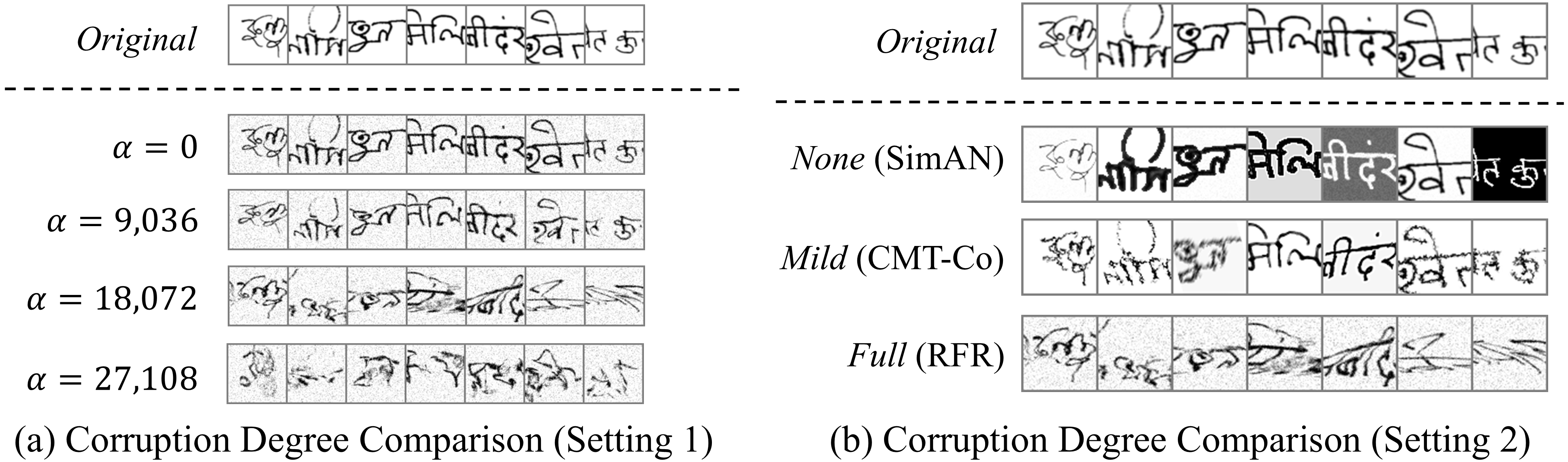}
	\caption{Visualized comparison of varying corruption degrees.}
	\label{Fig::vis_example}
\end{figure}

\subsection{Cross-Dataset Generalization Analysis}
\label{sec::cross-dataset}
To investigate cross-domain generalization of SigDiffuser, we conduct cross-dataset validation on four datasets, as shown in Table~\ref{Table::cross}. We measure performance degradation from source to target domain via accuracy drop ($\Delta_\text{Acc}$). All methods suffer noticeable degradation when transferred to unseen target datasets, with drops ranging from 3\% to 43\%. The decline is attributable to the domain-shift problem inherent in HSV~\cite{domainshift2020cvprw,domainshift2022ijcnn}, where variations in language, writing style, and data volume introduce significant discrepancies between datasets. Despite this challenging setting, SigDiffuser consistently yields smaller $\Delta_\text{Acc}$ than the baseline across different source-target pairs. For instance, when trained on BHSig260-H, it reduces the accuracy drops by 1.44\%, 5.85\%, and 4.56\% on BHSig260-B, CEDAR, and HanSig, respectively. It also achieves the smallest performance drop among compared methods in many settings, such as BHSig260-H$\rightarrow$BHSig260-B and HanSig$\rightarrow$CEDAR. These results indicate that SigDiffuser degrades more gracefully under domain shift than both the baseline and existing HSV methods.

Nevertheless, the absolute cross-dataset performance remains limited. SigDiffuser's accuracy falls below 80\% in 5 of the 12 cross-dataset settings, indicating a substantial gap from its in-domain performance (mostly above 92\%). This difficulty is not specific to our method: under the same protocol, previous SOTA methods such as ILNNeck and DetailSemNet exhibit comparable or more severe degradation, as evidenced by DetailSemNet falling below 80\% in 8 of the 12 settings. These observations suggest that cross-domain transfer in HSV remains an open problem, where inherent morphological differences between scripts introduce representation gaps that current methods, including ours, do not fully resolve.

Beyond script differences, cross-dataset generalization is further influenced by experimental factors. Following previous studies, we adopt the standard training/testing splits for each dataset, which results in varying numbers of training writers (\emph{e.g.}, 50 for CEDAR and 100 for BHSig260-H). This leads to asymmetric generalization behaviors, where models trained on larger splits naturally acquire richer representations. For instance, BHSig260-B and BHSig260-H share similar collection environments, but transferring from the larger BHSig260-H (100 training writers) to BHSig260-B (50 training writers) yields a much smaller performance drop than the reverse. This confirms that scaling up training writers helps mitigate the domain gap. This advantage is further magnified by writer diversity in HanSig. Results show that transferring to HanSig incurs larger drops than transferring from it to other datasets. This stems from its large training split (795 names, 31,800 samples), which exposes models to far more diverse writing styles, resulting in better cross-domain results.
%Together, these analyses indicate that cross-dataset results in HSV are jointly shaped by script differences and writer diversity across datasets, and methods explicitly designed for cross-script generalizability warrant dedicated future efforts.

\begin{figure*}[t]
	\centering
	\begin{subfigure}[t]{0.48\textwidth}
		\centering
		\includegraphics[width=0.98\linewidth]{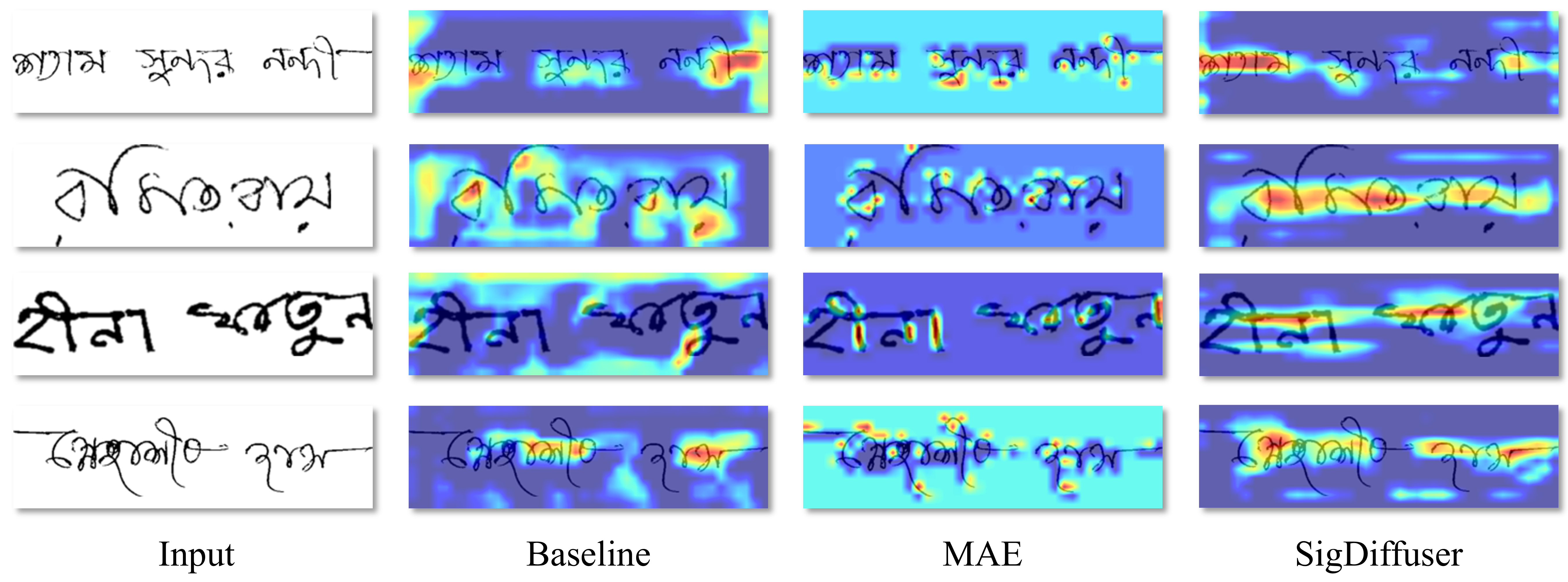}
		\caption{BHSig260-B.}
	\end{subfigure}
	\hfill
	\begin{subfigure}[t]{0.48\textwidth}
		\centering
		\includegraphics[width=0.98\linewidth]{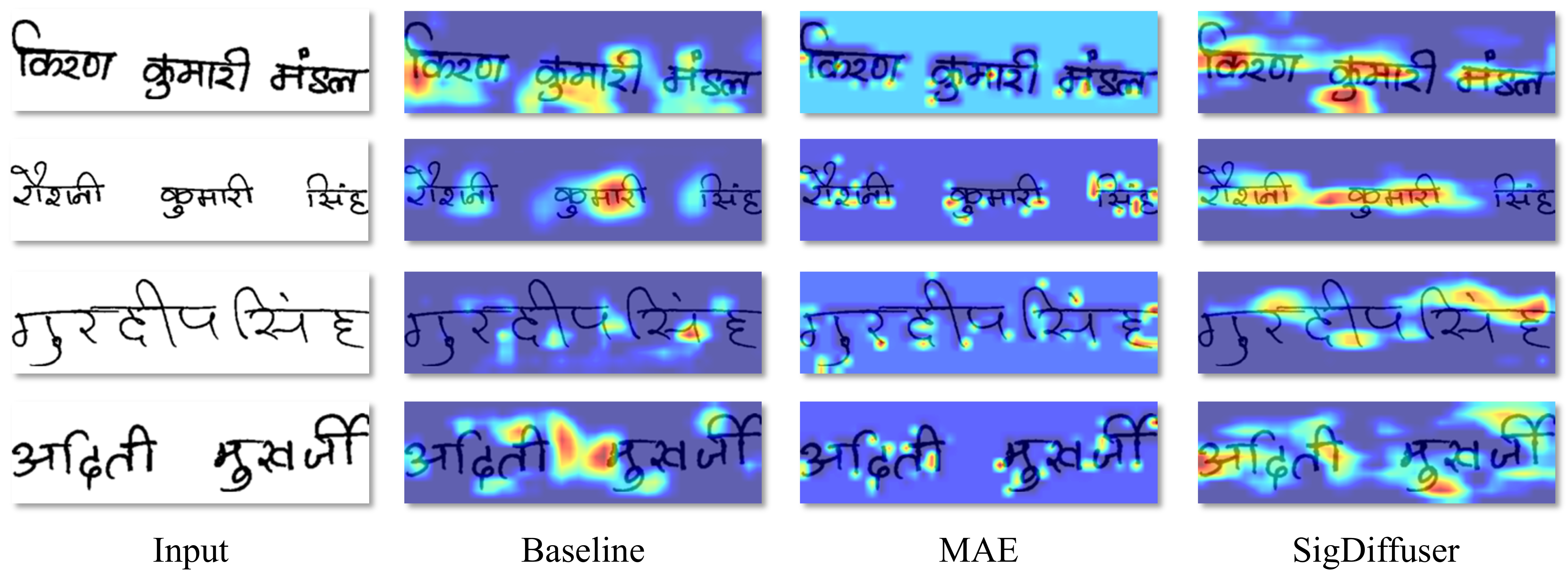}
		\caption{BHSig260-H.}
	\end{subfigure}
	\caption{Grad-CAM \cite{gradcam2017} visualizations of downstream models. While MAE produces sharp responses on local, isolated strokes, SigDiffuser highlights broader, discriminative structures like stroke connections and turns, capturing rich personal writing styles.}
	\label{Fig::gradcam}
\end{figure*}

\begin{figure*}[t]
	\centering
	\begin{subfigure}[t]{0.47\textwidth}
		\centering
		\includegraphics[width=0.95\linewidth]{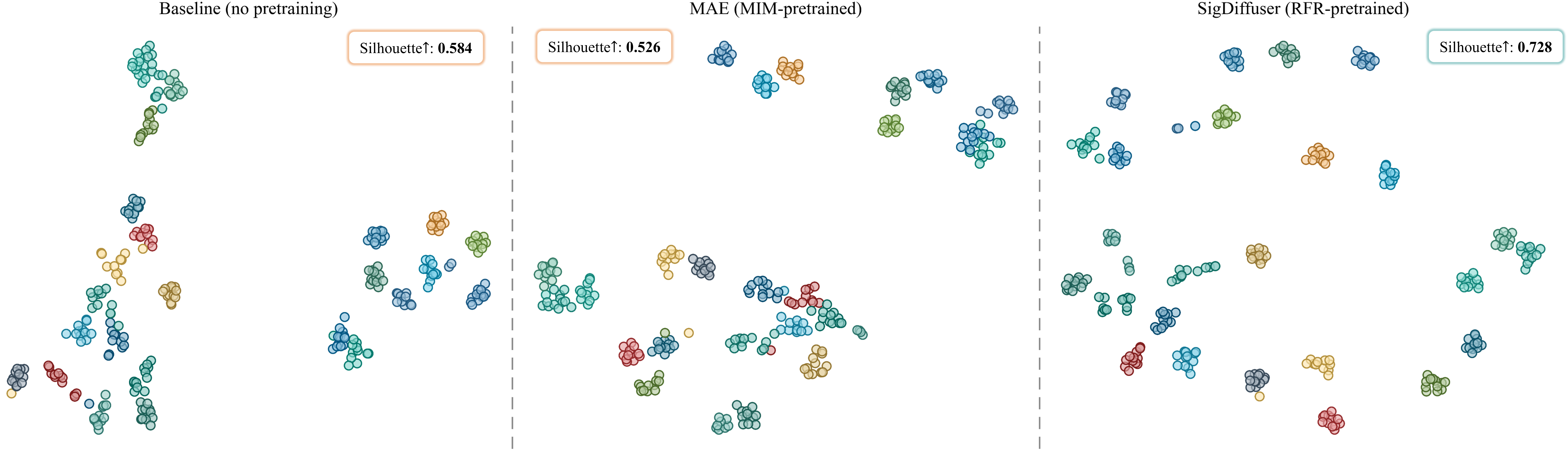}
		\caption{BHSig260-B: 12/24 samples per writer for 25/50 tested writers.}
	\end{subfigure}
	\hfill
	\begin{subfigure}[t]{0.47\textwidth}
		\centering
		\includegraphics[width=0.95\linewidth]{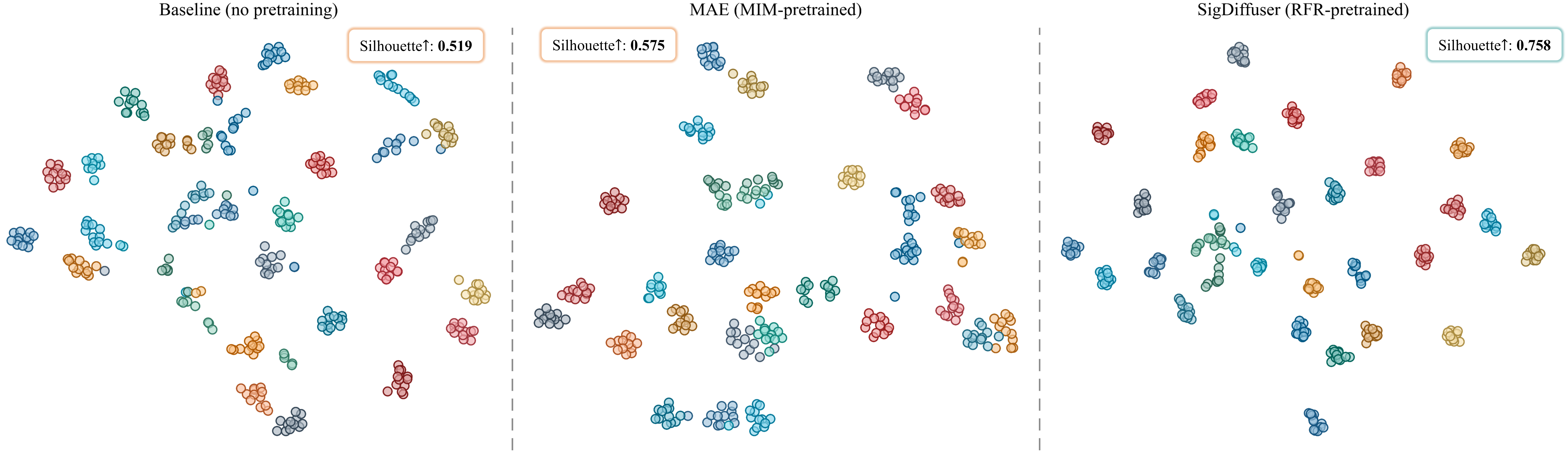}
		\caption{BHSig260-H. 12/24 samples per writer for 30/60 tested writers.}
	\end{subfigure}
	\caption{t-SNE visualizations of downstream HSV embeddings (partial samples shown; full results in Appendix H). We use the Silhouette score \cite{silhouettes} to quantify the separation degree (higher is better). Compared to the baseline and MAE, SigDiffuser forms more compact writer-level clusters with significantly higher scores, demonstrating superior feature separability.}
	\label{Fig::tsne}
\end{figure*}

\subsection{Effectiveness Analysis of RFR's Crumpling Strategy}
\label{sec::intep_crumple}
Given the unprecedentedly severe crumpling applied to signatures in RFR, we explore the effectiveness of crumpling degrees through two strategies. (1) Scale factor control. We vary $\alpha$ in Eq.~\ref{eq::alpha} across $0$, $9,036$, $18,072$, and $27,108$, representing moderate to severe crumpling. (2) Existing augmentation control. We apply data augmentation methods from SimAN \cite{siman2022luo} and CMT-Co \cite{cmtco2022zhang} on clean signatures to emulate no spatial distortion (\emph{None}) and mild crumpling (\emph{Mild}), with vanilla RFR crumpling as \emph{Full}. These proxies offer an alternative yet convincing evaluation of crumpling severity, with visualized and quantitative results shown in Fig.~\ref{Fig::vis_example} and Table~\ref{Table::comp_corrupt}.

Table~\ref{Table::comp_corrupt} (a) shows that performance improves as $\alpha$ increases, but excessive stroke warping ($\alpha=27,108$) exceeds reconstruction capacity and degrades results. Table~\ref{Table::comp_corrupt} (b) similarly shows that amplifying distortion progressively improves model performance. The consistency under both settings implies the effectiveness of introducing severe crumpling in RFR for empowering signature representation learning, which is a core difference against other generative methods that apply mild deformation \cite{siman2022luo} or simple noise addition \cite{diffae2022preechakul,ddae2023xiang,soda2024hudson}. In practice, one can tune $\alpha$ to find the optimal crumpling degree for application. Furthermore, we investigate employing forged signatures as natural crumpling in comparison to the artificial crumpling of RFR in Appendix G, which yields similar findings that reinforce RFR's efficacy.

\subsection{Visualization}
%\subsubsection{Feature Discriminability}
\label{sec::repr_vis}
We first evaluate feature discriminability using Grad-CAM \cite{gradcam2017} and t-SNE. Fig.~\ref{Fig::gradcam} shows Grad-CAM heatmaps using the ground-truth writer identity as targets. As observed, the baseline (no pretraining) generally fails to focus on valid signatures, while MAE (pretrained with MIM) sparsely activates on isolated stroke fragments, indicating shortcut learning based on superficial patterns. In contrast, SigDiffuser (pretrained with RFR) accurately attends to structurally significant regions like stroke connections and turns, which reflect intrinsic, unconscious personal writing habits. This confirms that the severe corruption design in RFR successfully suppresses direct reconstruction shortcuts, forcing the model to capture stable, discriminative writing-style evidence.

Subsequently, we present t-SNE visualizations of genuine signature features in Fig.~\ref{Fig::tsne} (partial samples shown for clarity; full results in Appendix H), illustrating how this attention translates into the global feature space. While baseline and MAE features exhibit noticeable class overlap and scattered distributions, SigDiffuser forms compact, well-separated writer clusters. This demonstrates that SigDiffuser effectively minimizes intra-writer variations and maximizes inter-writer separability, yielding discriminative representations. We further visualize SigDiffuser's reconstruction examples and failure cases in Appendices I and J.

\section{Limitation}
%\subsection{Technical Limitation}
\label{sec::challenge}
Apart from the strong HSV performance, some potential limitations warrant discussion. First, SigDiffuser primarily targets enhancing representation learning rather than image synthesis. Compared to dedicated image synthesis methods \cite{soda2024hudson,attribute2025tcsvt}, SigDiffuser may falter in producing a disentangled latent space and struggle with controllable image generation such as attribute manipulation. Second, while incorporating a broader range of signature data could further boost pretraining performance, we adhere to standard self-supervised protocols \cite{mae2022he,soda2024hudson} that restrict pretraining to the same dataset as fine-tuning to ensure fair comparisons. Third, while SigDiffuser achieves leading performance under in-domain settings (Tables~\ref{Table::finetune} to \ref{Table::comp_sota_utsig}), its absolute accuracy under cross-domain transfer (\emph{e.g.}, Bengali$\leftrightarrow$English$\leftrightarrow$Chinese) remains limited, as discussed in Sec.~\ref{sec::cross-dataset}. This is a field-wide challenge in HSV, and designing pretext tasks or domain-adaptation techniques that yield domain-invariant representations remains an important direction for future work.

\section{Conclusion}
\label{sec::conclusion}
We propose SigDiffuser, a novel diffusion-based representation learning method for handwritten signature verification (HSV), comprising a Reconstruction from Ruins (RFR) pretext task and a pyramidal feature aggregation module (PFAM). RFR introduces severe crumpling corruption on signatures, creating a challenging reconstruction problem that compels the model's deep understanding of discriminative writing patterns. PFAM aggregates multi-scale encoder features into latent embeddings, elevating representation quality.

Experiments across multiple HSV datasets demonstrate SigDiffuser's superior performance and adaptability across diverse linguistic and stylistic contexts. We reveal that increasing crumpling severity yields better signature representations, validating RFR's effectiveness. Unlike existing works \cite{diffae2022preechakul,soda2024hudson} that apply simple noise on clean images, our work innovatively proves that extreme destruction (\emph{i.e.}, severe crumpling) can enhance latent encoding for more informative representations. By leveraging denoising diffusion to reconstruct severely ``ruined'' signatures, we naturally bridge diffusion models and self-supervised learning, paving a new avenue for signature representation and potentially inspiring advances in other biometric domains, such as fingerprint verification.

\clearpage
\begin{appendices}

\section{Data Preprocessing}
\label{sec::preprocessing}
For data preprocessing, we first convert signature images into grayscale and remove the signature background with white pixel substitution. Gaussian filtering is then applied to filter out noise. Next, we excise the peripheral white pixel borders and resize the signature images to a fixed size of $96\times336$ (height $\times$ width), preserving the aspect ratio through white pixel padding until the preset dimensions are achieved to avoid distortions. Consequently, during pretraining, the patch side length $l$ is set to $96$, resulting in a size of $96\times96$ for cropped image patches $I_{org}$, $I_{cor}$, and $I_{sty}$. During downstream fine-tuning, we utilize $96\times336$ as the standard input size across all datasets.

\section{Per-Split Performance of SigDiffuser}
For transparency and reproducibility, we present the per-split performance of SigDiffuser on different datasets, along with comparisons against existing SOTA models. Results are presented in Tables~\ref{Table::per_split_bengali} to \ref{Table::per_split_cedar}. As observed, SigDiffuser surpasses previous SOTA in most single-split cases, especially on BHSig260-B, BHSig260-H, HanSig, and UTSig. This demonstrates that our model's superiority holds not only in the multi-split average but also across single splits. Yet, the EER on CEDAR fluctuates from 1.59\% (Split-1) to 18.55\% (Split-11), resulting in a suboptimal multi-split average. This sensitivity to data partitioning highlights an inherent limitation of traditional single-split protocols, where performance can be heavily biased by a specific random split. Therefore, it underscores the necessity of our multi-split evaluation for a more objective and reliable assessment.

\begin{table}[h]
	\renewcommand{\arraystretch}{1.1}
	\centering
	\caption{Per-split performance on BHSig260-B \cite{bhsig2016pal}.}
	\label{Table::per_split_bengali}
	\resizebox{\linewidth}{!}{
		\begin{tabular}{lcccc}
			\whline{1.05pt}
			Model & EER$\downarrow$ & FAR$\downarrow$ & FRR$\downarrow$ & Acc$\uparrow$ \\
			\hline
			\rowcolor{gray!8}
			\footnotesize \emph{Existing SOTA Model} & & & & \\
			CWA \cite{cwa2024ling} & 2.53 & 2.95 & 2.14 & 97.27 \\
			\rowcolor{gray!8}
			\footnotesize \emph{Ours} & & & & \\
			Split-1 & 1.59 & 1.59 & 1.59 & 98.41 \\
			Split-2 & 2.45 & 2.44 & 2.45 & 97.55 \\
			Split-3 & 2.23 & 2.23 & 2.22 & 97.77 \\
			Split-4 & 1.40 & 1.40 & 1.39 & 98.60 \\
			Split-5 & 1.58 & 1.58 & 1.57 & 98.43 \\
			\hline
			\noalign{\vspace{1pt}}
			Mean ($\pm$std) & \textbf{1.85} ($\pm$0.46) & \textbf{1.85} ($\pm$0.46) & \textbf{1.84} ($\pm$0.45) & \textbf{98.15} ($\pm$0.46) \\
			\whline{1.05pt}
	\end{tabular}}
\end{table}

\begin{table}[!t]
	\renewcommand{\arraystretch}{1.1}
	\centering
	\caption{Per-split performance on BHSig260-H \cite{bhsig2016pal}.}
	\label{Table::per_split_hindi}
	\resizebox{\linewidth}{!}{
		\begin{tabular}{lcccc}
			\whline{1.05pt}
			Model & EER$\downarrow$ & FAR$\downarrow$ & FRR$\downarrow$ & Acc$\uparrow$ \\
			\hline
			\rowcolor{gray!8}
			\footnotesize \emph{Existing SOTA Model}&&&&\\
			TransOSV \cite{transosv2024li} & 3.24 & 3.24 & 3.24 & - \\
			\rowcolor{gray!8}
			\footnotesize \emph{Ours} & & & & \\
			Split-1 & 3.13 & 3.13 & 3.13 & 96.87 \\
			Split-2 & 3.31 & 3.31 & 3.31 & 96.69 \\
			Split-3 & 3.44 & 3.44 & 3.44 & 96.56 \\
			Split-4 & 3.18 & 3.18 & 3.19 & 96.82 \\
			Split-5 & 2.59 & 2.59 & 2.58 & 97.41 \\
			\hline
			\noalign{\vspace{1pt}}
			Mean ($\pm$std) & \textbf{3.13} ($\pm$0.33) & \textbf{3.13} ($\pm$0.33) & \textbf{3.13} ($\pm$0.33) & \textbf{96.87} ($\pm$0.33) \\
			\whline{1.05pt}
	\end{tabular}}
\end{table}

\begin{table}[t]
	\renewcommand{\arraystretch}{1.1}
	\centering
	\caption{Per-split performance on HanSig \cite{hansig2023huang}.}
	\label{Table::per_split_hansig}
	\resizebox{\linewidth}{!}{
		\begin{tabular}{lcccc}
			\whline{1.05pt}
			Model & EER$\downarrow$ & FAR$\downarrow$ & FRR$\downarrow$ & Acc$\uparrow$ \\
			\hline
			\rowcolor{gray!8}
			\footnotesize \emph{Existing SOTA Model} & & & & \\
			MGRNet \cite{hansig2023huang} & 9.93 & 11.85 & 7.69 & - \\
			CWA \cite{cwa2024ling} & 9.47 & \textbf{6.61} & 16.23 & 90.26 \\
			\rowcolor{gray!8}
			\footnotesize \emph{Ours} & & & & \\
			Split-1 & 5.23 & 5.23 & 5.25 & 94.76 \\
			Split-2 & 7.36 & 7.36 & 7.36 & 92.64 \\
			Split-3 & 7.13 & 7.13 & 7.13 & 92.87 \\
			Split-4 & 5.81 & 5.81 & 5.81 & 94.19 \\
			Split-5 & 6.69 & 6.69 & 6.69 & 93.31 \\
			Split-6 & 6.84 & 6.84 & 6.83 & 93.16 \\
			Split-7 & 8.10 & 8.10 & 8.11 & 91.90 \\
			Split-8 & 6.83 & 6.82 & 6.83 & 93.18 \\
			Split-9 & 11.00 & 10.99 & 11.00 & 89.01 \\
			Split-10 & 8.89 & 8.89 & 8.89 & 91.11 \\
			\hline
			\noalign{\vspace{1pt}}
			Mean ($\pm$std) & \textbf{7.39} ($\pm$1.64) & 7.39 ($\pm$1.63) & \textbf{7.39} ($\pm$1.64) & \textbf{\textbf{92.61}} ($\pm$1.64)\\
			\whline{1.05pt}
	\end{tabular}}
\end{table}

\begin{table}[t]
	\renewcommand{\arraystretch}{1.1}
	\centering
	\caption{Per-split performance on UTSig \cite{utsig2017soleimani}.}
	\label{Table::per_split_utsig}
	\resizebox{\linewidth}{!}{
		\begin{tabular}{lcccc}
			\whline{1.05pt}
			Model & EER$\downarrow$ & FAR$\downarrow$ & FRR$\downarrow$ & Acc$\uparrow$ \\
			\hline
			\rowcolor{gray!8}
			\footnotesize \emph{Existing SOTA Model} & & & & \\
			UTSig-S1 \cite{utsig2017soleimani} & 29.71 & \textbf{21.29} & 39.27 & - \\
			HTCSigNet \cite{htcsignet2025zheng} & {28.93} & 28.92 & {28.96} & 71.08\\
			\rowcolor{gray!8}
			\footnotesize \emph{Ours} & & & & \\
			Split-1 & 24.22 & 24.22 & 24.23 & 75.78 \\
			Split-2 & 20.50 & 20.51 & 20.50 & 79.49 \\
			Split-3 & 25.18 & 25.17 & 25.19 & 74.83 \\
			Split-4 & 24.35 & 24.35 & 24.36 & 75.65 \\
			Split-5 & 23.14 & 23.14 & 23.15 & 76.86 \\
			\hline
			\noalign{\vspace{1pt}}
			Mean ($\pm$std) & \textbf{23.48} ($\pm$1.81) & 23.48 ($\pm$1.81) & \textbf{23.49} ($\pm$1.82) & \textbf{76.52} ($\pm$1.81) \\
			\whline{1.05pt}
	\end{tabular}}
\end{table}

\begin{table}[t]
	\renewcommand{\arraystretch}{1.1}
	\centering
	\caption{Per-split performance on CEDAR \cite{cedar2004kalera}.}
	\label{Table::per_split_cedar}
	\resizebox{\linewidth}{!}{
		\begin{tabular}{lcccc}
			\whline{1.05pt}
			Model & EER$\downarrow$ & FAR$\downarrow$ & FRR$\downarrow$ & Acc$\uparrow$ \\
			\hline
			\rowcolor{gray!8}
			\footnotesize \emph{Existing SOTA Model} & & & & \\
			IDN \cite{idn2019wei} & 3.62 & 5.87 & \textbf{2.17} & - \\
			SDINet \cite{sdinet2021li} & \textbf{1.75} & \textbf{0.73} & 3.42 & - \\
			\rowcolor{gray!8}
			\footnotesize \emph{Ours} & & & & \\
			Split-1 & 1.59 & 1.60 & 1.59 & 98.40 \\
			Split-2 & 3.26 & 3.26 & 3.26 & 96.74 \\
			Split-3 & 2.46 & 2.47 & 2.46 & 97.54 \\
			Split-4 & 2.25 & 2.26 & 2.25 & 97.75 \\
			Split-5 & 4.62 & 4.62 & 4.64 & 95.38 \\
			Split-6 & 5.02 & 5.03 & 5.00 & 94.98 \\
			Split-7 & 3.38 & 3.37 & 3.41 & 96.62 \\
			Split-8 & 8.06 & 8.06 & 8.04 & 91.95 \\
			Split-9 & 16.50 & 16.49 & 16.52 & 83.50 \\
			Split-10 & 10.42 & 10.42 & 10.43 & 89.58 \\
			Split-11 & 18.55 & 18.54 & 18.55 & 81.46 \\
			\hline
			\noalign{\vspace{1pt}}
			Mean ($\pm$std) & 6.92 ($\pm$5.88) & 6.92 ($\pm$5.87) & 6.92 ($\pm$5.88) & 93.08 ($\pm$5.88) \\
			\whline{1.05pt}
	\end{tabular}}
\end{table}

\section{Extended Ablation Study: Hierarchical Architecture Implementation}
To validate the effectiveness of the proposed pyramidal feature aggregation Module (PFAM) in hierarchical latent encoding, we conduct an ablation study regarding the implementation of multi-scale feature aggregation architectures. We implement three alternative hierarchical approaches to substitute the PFAM inside SigDiffuser, including channel-wise concatenation, cross-attention aggregation, and hierarchical bottlenecks. The first two directly integrate multi-scale features into decoder layers through concatenation or cross-attention. The hierarchical bottlenecks, which are an extension of the single bottleneck of SODA \cite{soda2024hudson}, compress multi-scale features into latent vectors and integrate them into decoder layers through addition. Table~\ref{Table::comp_arch} presents the comparative results. It can be seen that PFAM exhibits the optimal performance among the four methods, underscoring its superiority. By cumulatively compounding multi-scale handwriting feature maps, PFAM produces more informative latent representations than other architectures that linearly use multi-scale features without further aggregation. The enriched latent representations, in turn, enhance the encoder's feature extraction capabilities, leading to more effective representation learning.

\begin{table}[t]
	\renewcommand{\arraystretch}{1.15}
	\centering
	\caption{Comparison of different multi-scale feature aggregation architectures used for SigDiffuser on BHSig260-B \cite{bhsig2016pal} and BHSig260-H \cite{bhsig2016pal}. PFAM denotes the pyramidal feature aggregation module of SigDiffuser.}
	\label{Table::comp_arch}
	\resizebox{\linewidth}{!}{
		\begin{tabular}{c c c c c | c c c c}
			\toprule
			\multirow{2.4}{*}{Architecture} & \multicolumn{4}{c}{BHSig260-B} & \multicolumn{4}{c}{BHSig260-H}\\
			\cmidrule(r){2-5}\cmidrule(r){6-9}
			~ & EER $\downarrow$ & FAR $\downarrow$ & FRR $\downarrow$ & Acc $\uparrow$ & EER $\downarrow$ & FAR $\downarrow$ & FRR $\downarrow$ & Acc $\uparrow$\\
			\hline
			Concatenation & 1.77 & 1.77 & 1.77 & 98.23 & 3.83 & 3.83 & 3.83 & 96.17\\
			%			\hline
			Cross-Attention & 4.78 & 4.78 & 4.77 & 95.22 & 7.11 & 7.11 & 7.11 & 92.89\\
			%			\hline
			Hierarchical Bottleneck & 1.97 & 1.98 & 1.97 & 98.03 & \textbf{2.27} & \textbf{2.27} & \textbf{2.27} & \textbf{97.73}\\
			%			\hline
			\rowcolor{lightgray}
			PFAM & \textbf{1.58} & \textbf{1.58} & \textbf{1.57} & \textbf{98.43} & {2.59} & {2.59} & {2.59} & {97.41}\\
			\noalign{\vspace{-2pt}}
			\bottomrule
	\end{tabular}}
\end{table}

\begin{table}[t]
	\renewcommand{\arraystretch}{1.15}
	\centering
	\caption{Performance comparison of SigDiffuser pretrained with versus without diffusion modeling on BHSig260-B \cite{bhsig2016pal} and BHSig260-H \cite{bhsig2016pal}.}
	\label{Table::comp_diffusion}
	\resizebox{\linewidth}{!}{
		\begin{tabular}{c c c c c | c c c c}
			\toprule
			\multirow{2.4}{*}{Method} & \multicolumn{4}{c}{BHSig260-B} & \multicolumn{4}{c}{BHSig260-H}\\
			\cmidrule(r){2-5}\cmidrule(r){6-9}
			~ & EER $\downarrow$ & FAR $\downarrow$ & FRR $\downarrow$ & Acc $\uparrow$ & EER $\downarrow$ & FAR $\downarrow$ & FRR $\downarrow$ & Acc $\uparrow$\\
			\hline
			w/o Diffusion & 6.85 & 6.85 & 6.86 & 93.15 & 5.67 & 5.67 & 5.67 & 94.33\\
			%			\hline
			\rowcolor{lightgray}
			w/ Diffusion & \textbf{1.58} & \textbf{1.58} & \textbf{1.57} & \textbf{98.43} & \textbf{2.59} & \textbf{2.59} & \textbf{2.59} & \textbf{97.41}\\
			\noalign{\vspace{-2pt}}
			\bottomrule
	\end{tabular}}
\end{table}

\section{Extended Ablation Study: Effect of Diffusion Modeling in SigDiffuser}
In the RFR pretext task, SigDiffuser leverages the iterative denoising mechanism of diffusion to reconstruct severely crumpled signature patches for representation learning. To isolate the effect of the diffusion process, we remove it from the framework for an ablation. In this case, the framework degenerates into a single-step image-to-image translation model conditioned on an extra encoder. As shown in Table~\ref{Table::comp_diffusion}, performance drops significantly without the diffusion process. This is because restoring a heavily crumpled signature is far beyond the capability of a single-step mapping, which consequently hampers representation learning. In contrast, the multi-step iterative denoising of diffusion effectively handles such reconstruction, enabling the model to learn meaningful representations. These results highlight the indispensable role of diffusion modeling in our method, resonating with our motivation for coupling the RFR pretext task with a diffusion-based framework.

\section{Extended Ablation Study: Effect of Pretraining Epochs}
We further conduct an ablation study on the number of self-supervised pretraining epochs, with results shown in Table~\ref{Table::comp_epoch}. As observed, the performance improves as the pretraining schedule becomes longer, despite minor fluctuations. The best results are obtained at 800 epochs. This is particularly evident on BHSig260-H, where EER drops from 3.45\% to 2.59\% and Acc rises to 97.41\%.

These findings align with prior self-supervised learning methods, such as SimCLR~\cite{simclr2020icml}, MoCo v2~\cite{mocov22020}, SwAV~\cite{swav2020}, and MAE~\cite{mae2022he}, which demonstrate that longer pretraining generally improves downstream performance. Accordingly, they often adopt pretraining schedules of several hundred to over 1000 epochs, and inspire our adoption of 800 pretraining epochs. Although extended pretraining schedules are often studied on large-scale datasets such as ImageNet \cite{imagenet2009deng}, our results indicate that this practice also benefits the proposed RFR task on relatively smaller signature datasets. Therefore, we adopt 800 epochs as the default pretraining schedule.

\begin{table}[t]
	\renewcommand{\arraystretch}{1.1}
	\centering
	\caption{Performance comparison under different numbers of pretraining epochs in the RFR task on BHSig260-B \cite{bhsig2016pal} and BHSig260-H \cite{bhsig2016pal}.}
	\label{Table::comp_epoch}
	\resizebox{\linewidth}{!}{
		\begin{tabular}{c c c c c | c c c c}
			\toprule
			\multirow{2.4}{*}{\#Epoch} & \multicolumn{4}{c}{BHSig260-B} & \multicolumn{4}{c}{BHSig260-H}\\
			\cmidrule(r){2-5}\cmidrule(r){6-9}
			~ & EER $\downarrow$ & FAR $\downarrow$ & FRR $\downarrow$ & Acc $\uparrow$ & EER $\downarrow$ & FAR $\downarrow$ & FRR $\downarrow$ & Acc $\uparrow$\\
			\hline
			300 & 2.13 & 2.13 & 2.14 & 97.87 & 3.19 & 3.19 & 3.19 & 96.81\\
			%			\hline
			500 & 1.90 & 1.90 & 1.90 & 98.10 & 3.21 & 3.21 & 3.21 & 96.79\\
			%			\hline
			700 & 2.14 & 2.13 & 2.14 & 97.86 & 2.63 & 2.63 & 2.63 & 97.37\\
			%			\hline
			\rowcolor{lightgray}
			800 & \textbf{1.58} & \textbf{1.58} & \textbf{1.57} & \textbf{98.43} & \textbf{2.59} & \textbf{2.59} & \textbf{2.59} & \textbf{97.41}\\
			\noalign{\vspace{-2pt}}
			\bottomrule
	\end{tabular}}
\end{table}

\section{Extended Ablation Study: Effect of Patch Side Length}
\label{sec::sidelength}
The side length $l$ of the signature square patch used in RFR could essentially affect model performance. Consequently, we conduct ablation studies on the side length, with results presented in Table~\ref{Table::comp_patchl}. From the outcomes, increasing $l$ steadily improves the downstream performance, achieving the optimal results at $l=96$. Further increasing $l$ to 128 leads to an obvious drop in performance. The optimal value of $l=96$ naturally aligns with our downstream evaluation setup, which uses input dimensions of $96\times336$, ensuring better model perception of signature pixel distributions. Based on these results, we adopt $l=96$ for all experiments.

\begin{figure*}[t]
	\centering
	\begin{subfigure}[t]{0.48\textwidth}
		\centering
		\includegraphics[width=\linewidth]{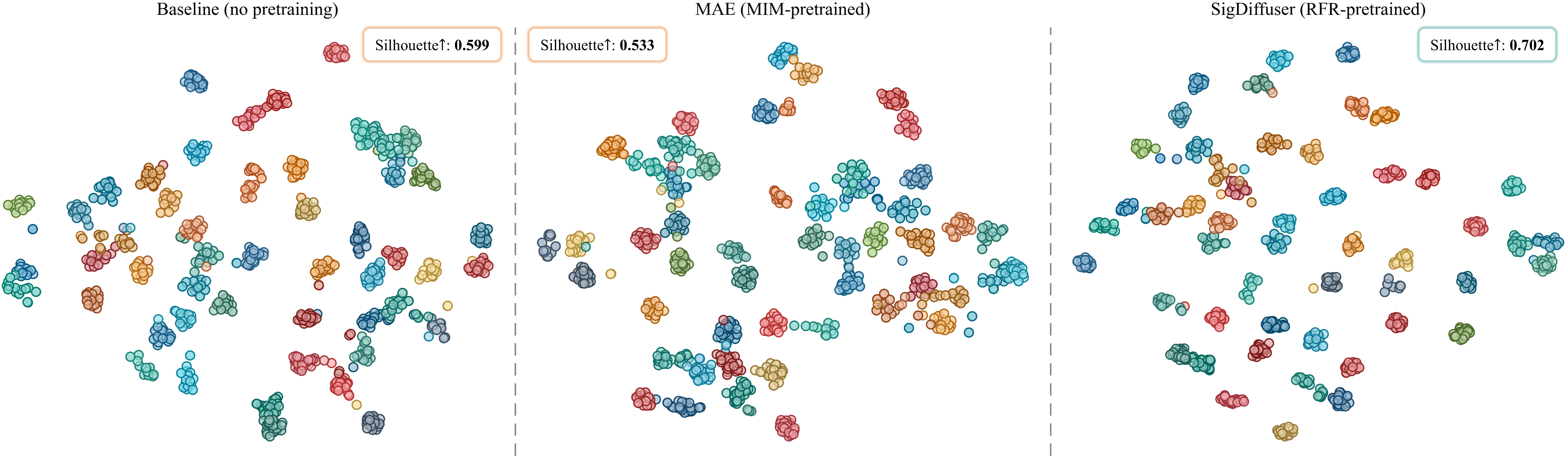}
		\caption{BHSig260-B.}
	\end{subfigure}
	\hfill
	\begin{subfigure}[t]{0.48\textwidth}
		\centering
		\includegraphics[width=\linewidth]{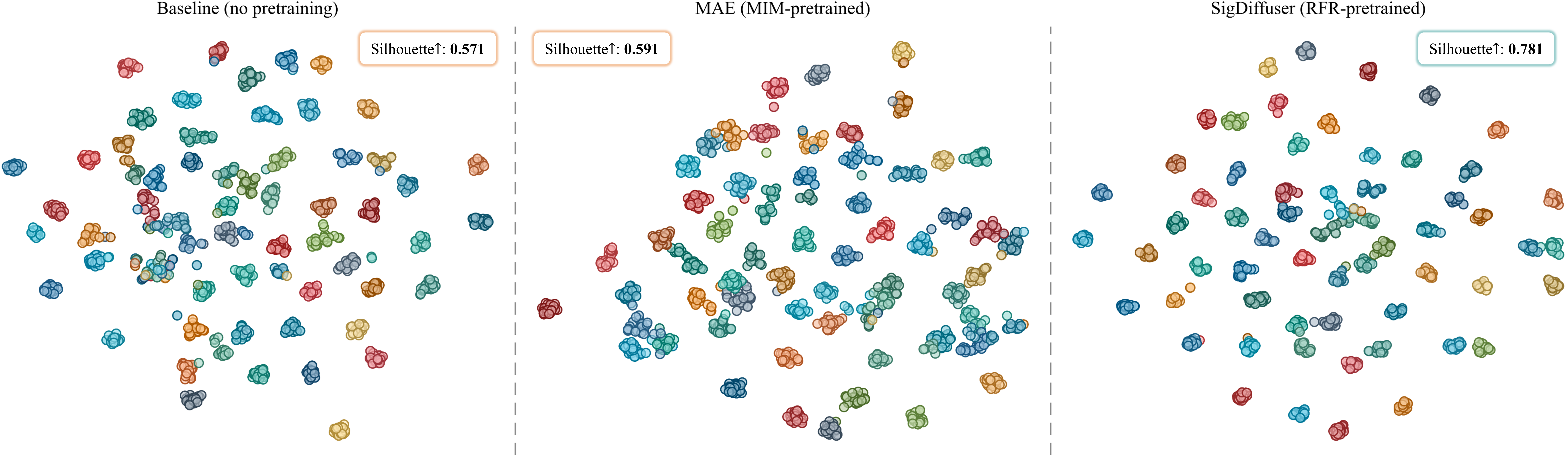}
		\caption{BHSig260-H.}
	\end{subfigure}
	\caption{Full t-SNE visualizations \cite{tsne2008} of downstream HSV embeddings. We use the Silhouette score \cite{silhouettes} to quantify the compactness and separation (higher is better). SigDiffuser yields more compact writer-level feature clusters with significantly larger Silhouette scores than the baseline and MAE, validating its effectiveness in learning separable writer-level representations.}
	\label{Fig::tsne_full}
\end{figure*}

\begin{table}[t]
	\renewcommand{\arraystretch}{1.1}
	\centering
	\caption{Performance comparison of varying signature patch size (side length) $l$ used in the RFR task on BHSig260-B \cite{bhsig2016pal} and BHSig260-H \cite{bhsig2016pal}.}
	\label{Table::comp_patchl}
	\resizebox{\linewidth}{!}{
		\begin{tabular}{c c c c c | c c c c}
			\toprule
			\multirow{2.4}{*}{$l$} & \multicolumn{4}{c}{BHSig260-B} & \multicolumn{4}{c}{BHSig260-H}\\
			\cmidrule(r){2-5}\cmidrule(r){6-9}
			~ & EER $\downarrow$ & FAR $\downarrow$ & FRR $\downarrow$ & Acc $\uparrow$ & EER $\downarrow$ & FAR $\downarrow$ & FRR $\downarrow$ & Acc $\uparrow$\\
			\hline
			32 & 4.92 & 4.92 & 4.91 & 95.08 & 4.44 & 4.44 & 4.44 & 95.56\\
			%			\hline
			48 & 3.07 & 3.08 & 3.07 & 96.93 & 4.69 & 4.70 & 4.69 & 95.30\\
			%			\hline
			64 & 2.38 & 2.38 & 2.38 & 97.62 & 3.38 & 3.38 & 3.38 & 96.62\\
			%			\hline
			\rowcolor{lightgray}
			96 & \textbf{1.58} & \textbf{1.58} & \textbf{1.57} & \textbf{98.43} & \textbf{2.59} & \textbf{2.59} & \textbf{2.59} & \textbf{97.41}\\
			%			\hline
			128 & 1.66 & 1.67 & 1.66 & 98.34 & 4.02 & 4.02 & 4.03 & 95.98\\
			\noalign{\vspace{-2pt}}
			\bottomrule
	\end{tabular}}
\end{table}

\begin{table}[t]
	\renewcommand{\arraystretch}{1.2}
	\centering
	\caption{Comparison between using forged signatures (natural destruction) and RFR's severe crumpling as destruction methods on BHSig260-B \cite{bhsig2016pal} and BHSig260-H \cite{bhsig2016pal}.}
	\label{Table::comp_gf}
	\resizebox{\linewidth}{!}{
		\begin{tabular}{c c c c c | c c c c}
			\toprule
			\multirow{2.4}{*}{Crumpling Effect} & \multicolumn{4}{c}{BHSig260-B} & \multicolumn{4}{c}{BHSig260-H}\\
			\cmidrule(r){2-5}\cmidrule(r){6-9}
			~ & EER $\downarrow$ & FAR $\downarrow$ & FRR $\downarrow$ & Acc $\uparrow$ & EER $\downarrow$ & FAR $\downarrow$ & FRR $\downarrow$ & Acc $\uparrow$\\
			\hline
			Forged Signature & 3.04 & 3.04 & 3.04 & 96.96 & 4.09 & 4.09 & 4.09 & 95.91\\
			%			\hline
			\rowcolor{lightgray}
			Vanilla Crumpling & \textbf{1.58} & \textbf{1.58} & \textbf{1.57} & \textbf{98.43} & \textbf{2.59} & \textbf{2.59} & \textbf{2.59} & \textbf{97.41}\\
			\noalign{\vspace{-2pt}}
			\bottomrule
	\end{tabular}}
\end{table}

\section{Effectiveness Analysis of RFR's Crumpling Strategy: Using Forgery as Natural Crumpling}
While RFR employs artificial crumpling, skillfully forged signatures represent natural deformations \cite{pavenet2025zhang} of genuine signatures. Therefore, we compare using forgeries as corrupted images in RFR against the vanilla crumpling, with results presented in Table~\ref{Table::comp_gf}. As observed, downstream performance decreases when forged signatures are used as crumpling. The reasons could be two-fold. (1) Using forged signatures as corruptions reduces the pretraining data volume by over half, which could decline downstream performance. (2) The natural variations of forged signatures exhibit fewer spatial deformations compared to vanilla corrupted ones, weakening the model's representation learning ability. This aligns with the finding in Sec.~IV-G that a lower corruption degree declines model performance, reinforcing the efficacy of using severe crumpling in RFR.

\section{Full t-SNE Visualization}
Complementing Sec. IV-H1 in the main paper, we present the full t-SNE visualization results in Fig.~\ref{Fig::tsne_full}. Compared with the baseline model and MAE, SigDiffuser forms significantly more compact and better separated writer-level clusters, with higher Silhouette scores. The visualized and numerical results collectively validate SigDiffuser's effectiveness in learning discriminative signature representations.

\begin{figure*}[t]
	\centering
	\includegraphics[width=0.96\textwidth]{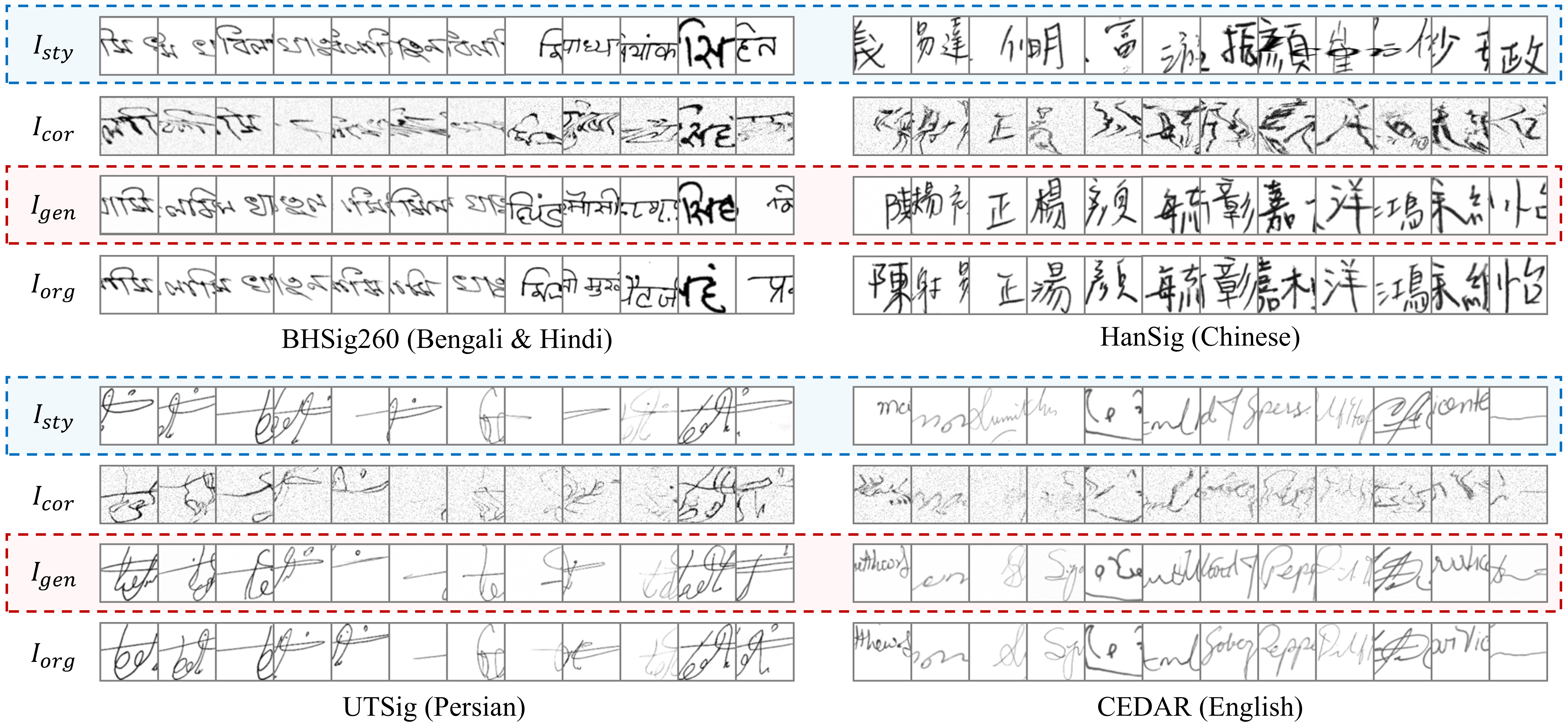}
	\caption{Reconstruction visualizations of SigDiffuser on different datasets. $I_{sty}$ denotes the intact signature patches used as the style reference, which is input to the visual encoder. $I_{cor}$ denotes the corrupted signature patch. $I_{org}$ denotes the clean version of $I_{cor}$, which is also the reconstruction target of the model. $I_{gen}$ denotes the reconstructed signature patches.}
	\label{Fig::visualize}
\end{figure*}

\begin{figure*}[t]
	\centering
	\includegraphics[width=\textwidth]{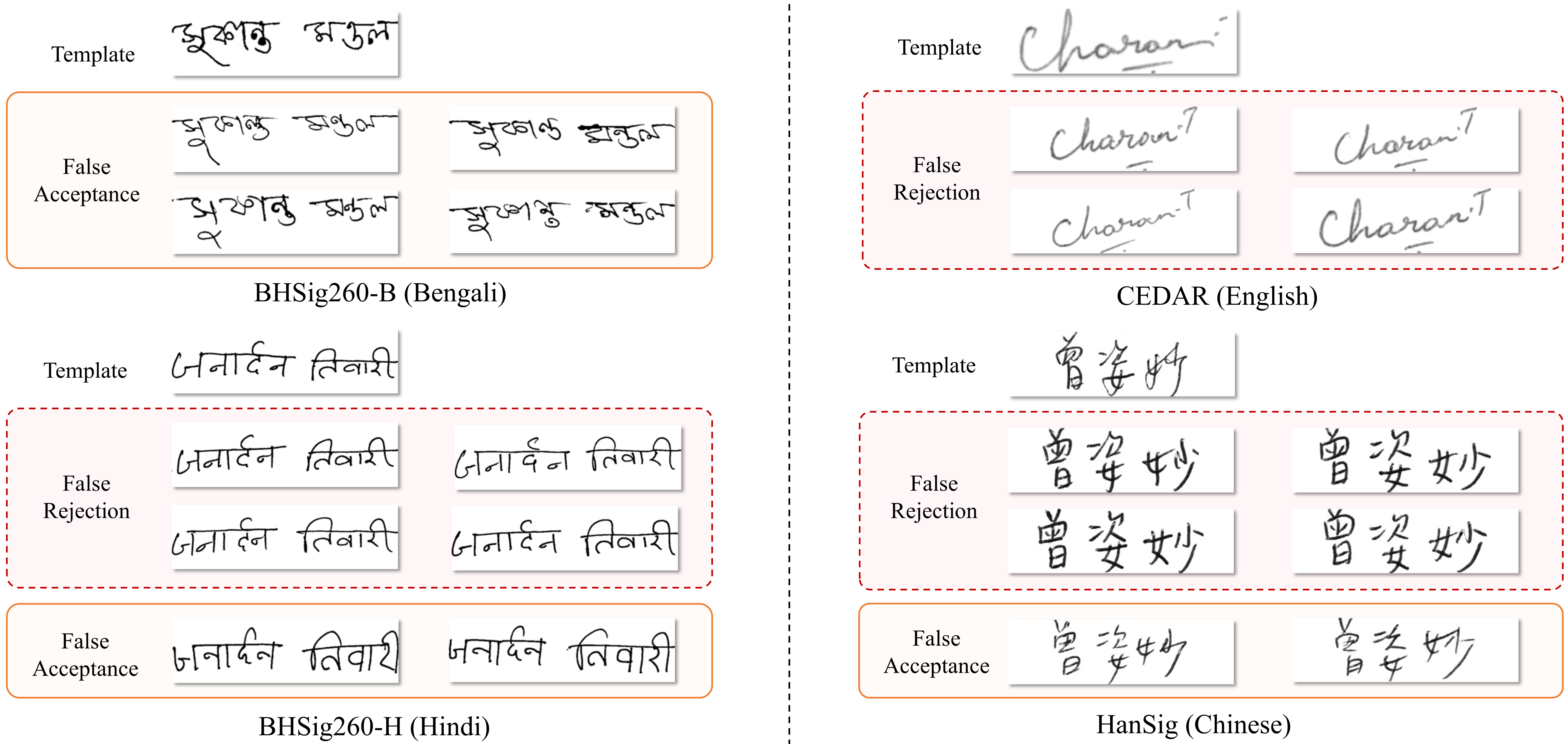}
	\caption{Visualization of false rejection and false acceptance examples of SigDiffuser.}
	\label{Fig::false}
\end{figure*}

\section{Reconstruction Example}
\label{sec::vis}
We present reconstruction examples across datasets in Fig.~\ref{Fig::visualize}. Comparing the reconstructed signatures $I_{gen}$ to the original uncrumpled ones $I_{org}$, writing styles and textures are faithfully restored, while the content is sometimes imprecisely replicated as visually plausible but garbled characters. Nonetheless, unlike traditional font/text generation methods \cite{fontdiffuser2024yang,slogan2023luo}, the primary objective of RFR is on capturing signature styles rather than precise verbatim replication. Despite these discrepancies, SigDiffuser achieves high-quality signature representations through self-supervised pretraining, as demonstrated across various datasets (Tables~II to VI in the main paper). Furthermore, as shown in Fig.~1, SigDiffuser's reconstructions significantly surpass those of Masked Image Modeling (MIM), which only restores blurred structures. This superior structural fidelity demonstrates a deep understanding of signature writing patterns, confirming that precise content replication is inconsequential to our method's overall effectiveness as an HSV pretraining approach.

\section{Failure Case Analysis}
We present some incorrectly verified samples of SigDiffuser across different datasets in Fig.~\ref{Fig::false}, including false rejection and false acceptance. As observed, the false rejections typically exhibit significant deviations from their reference templates. For example, in CEDAR, the falsely rejected samples show larger inclination angles and contrasting size proportions (shorter overall length) compared to the template. In HanSig, the template signature displays a free-flowing writing style with obvious connected strokes, whereas the false rejection lacks stroke connections and exhibits a rigid writing style. The huge differences lead to their misclassification as forgeries, despite originating from the same writers of templates.

Regarding false acceptances, forged signatures often exhibit local writing patterns that resemble their corresponding template. For instance, the central portion of Bengali forgeries, the stroke inclination in Hindi forgeries, and the free-flowing writing style in Chinese forgeries all demonstrate similarities to their respective templates. The model may not always identify these deceptive details, leading to erroneous classification. These failed cases reveal that self-supervised learning remains challenged by some high-quality forged samples and struggles with genuine samples that exhibit significant stylistic variations from templates, despite improved HSV performance. Notably, Tables III to VII show that SigDiffuser maintains closely balanced FAR and FRR, suggesting the need for comprehensive optimization across all failure modes rather than targeting specific error types.
\end{appendices}

\bibliographystyle{IEEEtran}
\bibliography{reference}

\end{document}